\documentclass[review]{elsarticle}

\usepackage{amssymb, comment}
\usepackage{mathtools}
\usepackage{breqn}
\usepackage{algorithm}
\usepackage{multirow}
\usepackage[noend]{algpseudocode}
\usepackage{pgfplots} 
\pgfplotsset{compat=1.9}
\usepackage{collcell}
\usepackage{tikz}
\usepackage{natbib}
\usetikzlibrary{patterns}
\usetikzlibrary{decorations.pathreplacing}
\usepackage{soul}
\usepackage{subfigure}
\usepackage{adjustbox}
\usetikzlibrary{shapes.geometric, arrows, fit}
\usepackage{lineno,hyperref}
\modulolinenumbers[5]
\usepackage{makecell}

\newcommand{\CC}{C\nolinebreak\hspace{-.05em}\raisebox{.4ex}{\tiny\bf +}\nolinebreak\hspace{-.10em}\raisebox{.4ex}{\tiny\bf +}}
\def\CC{{C\nolinebreak[4]\hspace{-.05em}\raisebox{.4ex}{\tiny\bf ++ }}}

\newsavebox\DontUseBreqn

\def\F{\mathbf{F} }

\def\Proof{\noindent{\sl Proof.}\ }
\def\qed{{\hfill $\Box$ \medbreak}}
\newcolumntype{P}[1]{>{\centering\arraybackslash}p{#1}}

\tikzstyle{startstop} = [rectangle, rounded corners, minimum width=2cm, minimum height=1cm,text centered, draw=black]
\tikzstyle{io} = [trapezium, trapezium left angle=70, trapezium right angle=110, minimum width=1cm, minimum height=1cm, text centered,  draw=black]
\tikzstyle{process} = [rectangle, minimum width=3cm, minimum height=1cm, text centered, text width=3.8cm, draw=black]
\tikzstyle{process1} = [rectangle, minimum width=5cm, minimum height=1cm, text centered, text width=6.8cm, draw=black]
\tikzstyle{decision} = [diamond, minimum width=3cm, minimum height=1cm, text centered, draw=black]
\tikzstyle{arrow} = [thick,->,>=stealth]
\tikzset{container/.style={draw, rectangle, dashed, inner sep=.3em}}
\tikzset{container1/.style={draw, rectangle, dashed, inner sep=.6em}}

\newtheorem{defi}{Definition}[section]
\newtheorem{thm}[defi]{Theorem}

\newcommand{\textbox}[1]
{\savebox{\tempbox}{#1}
	\ifdim\wd\tempbox<4cm\relax
	\makebox[4cm]{\usebox{\tempbox}}%
	\else
	\parbox{4cm}{\raggedright #1}%
	\fi}

\tikzset{
	max width/.style args={#1}{
		execute at begin node={\begin{varwidth}{#1}},
			execute at end node={\end{varwidth}}
	}
}

\journal{Knowledge-Based Systems}

\begin{document}
	
	\begin{frontmatter}
		
		
		
		\title{High-Dimensional Feature Selection for Genomic Datasets}
		
		\author[label1]{Majid Afshar}
		\address[label1]{Department of Computer Science, Memorial University of Newfoundland\\ St. John's, NL, Canada}
		\author[label2]{Hamid Usefi}
		\address[label2]{Department of Mathematics and Statistics, Memorial University of Newfoundland\\ St. John's, NL, Canada}
		
		\begin{abstract}
			A central problem in machine learning and pattern recognition is the process of recognizing the most important features.  In this paper, we provide a new feature selection method (DRPT) that consists of first removing the irrelevant features and then detecting correlations between the remaining features. 
			Let  $D=[A\mid \mathbf{b}]$ be a dataset, where $\mathbf{b}$ is the class label and $A$ is a matrix whose columns are  the features. We solve $A\mathbf{x} = \mathbf{b}$   using  the  least squares method and the pseudo-inverse of $A$.
			 Each component of $\mathbf{x}$ can be viewed as an assigned weight to the corresponding column (feature). We define a threshold based on the local maxima of $\mathbf{x}$ and remove those features whose weights are smaller than the threshold.
			 To detect the correlations in the reduced matrix, which we still call $A$, we consider a perturbation $\tilde A$ of $A$. We prove that correlations are encoded in
			  $\Delta\mathbf{x}=\mid \mathbf{x} -\tilde{\mathbf{x}}\mid $, where $\tilde{\mathbf{x}}$ is the least quares solution of 
			  $\tilde A\tilde{\mathbf{x}}=\mathbf{b}$. We cluster features first based on $\Delta\mathbf{x}$ and then using the entropy of features. Finally, a feature is  selected from each sub-cluster based on its weight and entropy. 
			The effectiveness of DRPT has been verified by performing a series of comparisons with seven state-of-the-art feature selection methods over ten genetic datasets ranging up from 9,117 to 267,604 features.  The results show that, over all,  the performance of DRPT  is favorable in several aspects compared to each feature selection algorithm. 
		\end{abstract}
		
		
	\end{frontmatter}
	
	
	\section{Introduction}
	Supervised learning is a central problem in machine learning and data mining \cite{breiman2017classification}. In this process, a mathematical/statistical model is trained and generated based on a pre-defined number of instances (train data) and is tested against the remaining (test data). A subcategory of supervised learning is classification, where the model is trained to predict class labels \cite{nasrabadi2007pattern}. For instance, in tumor datasets, class labels can be malignant or benign, the former being cancerous and the latter being non-cancerous tumors \cite{sorlie2001gene}. For each instance in a classification problem, there exists a set of features that contribute to the output \cite{john1994irrelevant}.
	
	In high-dimensional datasets, there are a large number of  irrelevant features that have no correlation with the class labels.  Irrelevant features act as noise in the data that not only  increase the  computational costs but, in some cases, divert the learning process toward weak model generation \cite{khalid2014survey, ding2005minimum}. The other important  issue   is the presence of correlation between good features, which makes some features redundant.  Redundancy is  known as  multicollinearity in a broader context  and it  is known to create  overfitting and bias in regression when 
 a model is trained on data from one region and predicted on another region  \cite{Collinearity, tamura2017best}.

	The goal of feature selection (FS) methods is to select the most important and effective features \cite{guyon2003introduction}. As such, FS can decrease the model complexity in the training phase while  retaining or  improving the classification accuracy. 
	Recent FS methods \cite{gaudioso2017lagrangian,Xue2018,10.1007/978-981-10-7566-7_22} usually find the most important features through a complex model which introduce a more complicated framework when followed by a classifier.

	In this paper, we present a linear FS method called dimension reduction based on perturbation theory (DRPT).  Let  $D=[A\mid \mathbf{b}]$ be a dataset where $\mathbf{b}$ is the class label and $A$ is an $m\times n$ matrix whose columns are  features. We shall focus on datasets where  $m<<n$ and of particular interests to us are genomic datasets where gene expression level of  samples (cases and controls) are measured. So, each feature is the expression levels of a gene measured across all samples. Biologically speaking, there is only a limited number of genes that are associated to a disease and, as such, only expression levels of certain genes can differentiate between cases and controls. So, a majority of genes are considered irrelevant. 
	One of the most common methods to   filter out irrelevant features in genomic datasets is using $p$-values. That is, one can look at the expression levels of a gene in normal and disease cases and   calculate the $p$-values based on some statistical tests. It has been customary to conclude that genes whose $p$-values are not significant are irrelevant and  can be filtered out.   However,  genes expressions are not independent events (variables) and researchers have been  warned against the misuse of statistical significance and $p$-values,  as it is recently pointed out in \cite{amrhein2019scientists, wasserstein2019moving}. 
	
	We consider the system $A\mathbf{x}=\mathbf{b}$ where   the rows of $A$ are independent of each other and   $A\mathbf{x}=\mathbf{b}$ is an underdetermined linear system. This is the case for genomic datasets because each sample has different gene expressions from the others. Since $A\mathbf{x} = \mathbf{b}$ may not have a unique solution, instead we  use  the  least squares method and the pseudo-inverse of $A$ to find the solution  with the smallest 2-norm.  One can view each component $x_i$  of $\mathbf{x}$ as an assigned weight to the column (feature)  $\mathbf{F}_i$ of $A$. Therefore,  the bigger the $|x_i|$ the more important  $\mathbf{F}_i$ is in connection with  $\mathbf{b}$. 
	
	It then makes sense to filter out those features whose weights are very small compared to the average of local maximums over  $|x_i|$'s.  After removing irrelevant features, we obtain a reduced dataset,  which we still denote  it by $[A\mid \mathbf{b}]$. In the next phase, we detect correlations between columns of $A$ by perturbing $A$ using a randomly generated  matrix  $E$    of small norm. Let $\tilde{ \mathbf{x}}$ be the solution to $(A+E)\tilde{\mathbf{x}}=\mathbf{b}$. It follows from Theorem \ref{perturb} that features $\mathbf{F}_i$ and $\mathbf{F}_j$ correlate if and only if $| x_i-\tilde x_i|$ and $| x_j-\tilde x_j|$ are almost the same.
	Next, we cluster  $\Delta \mathbf{x}= |\mathbf{x}-\tilde{\mathbf{x}}|$  using a simplified least-squares method called Savitsky-Golay  smoothing filter \cite{schafer2011savitzky}.  This process yields a step-wise function where each step is a cluster.  We  note that  features in the same cluster do not necessarily correlate and so  we further break up each cluster of  $\Delta \mathbf{x}$ into sub-clusters using   entropy of features. Finally, from each sub-cluster,  we pick a feature and rank all the selected features using entropy.
	
	The ``stability'' of a feature selection algorithm is recently discussed in \cite{nogueira2017stability}.  An algorithm is `unstable' if a small change in data leads to large changes in the chosen feature subset.
	In real datasets, it is possible that there are small noise or error involved in the data.  Also, the order of samples (rows) in a dataset should not matter; the same applies to the order of features (columns). In Theorem \ref{noise} we prove that DRPT is noise-robust and in Theorem \ref{shuffle} we prove that DRPT is stable with respect to permuting rows or columns.

	We compare our method with seven state-of-the-art FS methods, namely 
	mRMR \cite{peng2005feature},  LARS \cite{efron2004least},  HSIC-Lasso \cite{yamada2014high}, 
	Fast-OSFS \cite{wu2012online},  group-SAOLA  \cite{yu2016scalable}, CCM \cite{chen2017kernel} and BCOA \cite{de2020binary} 
	over ten genomic datasets ranging up from 9,117 to 267,604 features. We use Support Vector Machines (SVM) and Random Forest (RF) to classify the datasets based on the selected features by FS algorithms. 	The results show that, over all,  the classification accuracy of DRPT  is favorable  compared with   each individual FS algorithm. Also, we report the running time, CPU time, and memory usage and  demonstrate that DRPT does well compared to other  FS methods.

	The rest of this paper is organized as follows. In Section \ref{related}, we review related work. 
	We present our approach and the algorithm in Section \ref{approach}. Experimental results and performance comparisons are shown in Section \ref{Results} and we conclude the paper in Section \ref{Conclusions}.
	
\section{Related Work}\label{related}
	FS methods are  categorized as filter, wrapper and embedded methods \cite{kohavi1997wrappers}. Filter methods evaluate each feature regardless of the learning model. Wrapper-based methods select features by  assessing the prediction power of each feature provided by a classifier. The quality of the selected subset using these methods is very high, but  wrapper methods are computationally inefficient. The last group consolidates the advantages of both methods, where a given classifier selects the most important features simultaneously with the training phase. These methods are powerful, but the feature selection process cannot be defused from the classification process. 
	
	Providing the most informative and important features to a classifier would result in a better prediction power and higher accuracy \cite{yang1997comparative}. Selecting an optimal subset of features  is an NP-hard problem that  has attracted many researchers to apply meta-heuristic and stochastic algorithms \cite{xue2016survey,wang2015multi,paul2015simultaneous}.

	Several methods exist that aim to enhance classification accuracy by assigning a common discriminative feature set to local behavior of data in different regions of the feature space \cite{sun2009local,armanfard2015local}. For example, localized feature selection (LFS) is introduced by Armanfard et al. \cite{armanfard2015local}, in which a set of features is selected to accommodate a subset of samples. For an arbitrary query of the unseen sample, the similarity of the sample to the representative sample of each region is calculated, and the class label of the most similar region is assigned to the new sample.

	On the other hand, some  approaches use aggregated sample data to select and rank the features  \cite{peng2005feature, efron2004least,yamada2014high,tibshirani1996regression,chen2017semi}. The least absolute shrinkage and selection operator (LASSO) is an estimation method in linear models which simultaneously applies variable selection by setting some coefficients to zero \cite{tibshirani1996regression}. 

	Least angle regression (LARS) proposed by Efron et al. \cite{efron2004least} is based on LASSO and is a linear regression method which computes all least absolute shrinkage and selection operator \cite{tibshirani1996regression} estimates and selects those features which are highly correlated to the already selected ones. 
	
	Chen et al. \cite{chen2017semi} introduced a semi-supervised FS called Rescaled Linear Square Regression
	(RLSR), in which rescaling factors are incorporated to exploit the least square regression model and rank features. They solve the minimization problem shown in equation \ref{RLSR} to learn $\mathbf{\Theta}$ and $\mathbf{Y}_U$, which are a matrix of rescaling factors and unknown labels, respectively.
	
	\begin{dmath}\label{RLSR}
		\min \Big (||\mathbf{X}^T \mathbf{\Theta} \mathbf{W} + \mathbf{1b}^T - \mathbf{Y}||^2_F + \gamma ||\mathbf{W}||^2_F \Big )\\
		\text{st.} \mathbf{W},\mathbf{b}, \theta > 0, \mathbf{1}^T \theta=1, \mathbf{Y}_U \ge 0, \mathbf{Y}_U\mathbf{1}=\mathbf{1},
	\end{dmath}
	where $\mathbf{W}$ is a sparse matrix that represents the importance of features, $\mathbf{X}$ is the dataset, $\mathbf{Y}$ is class labels, and $b=\frac{1}{n}(\mathbf{Y}^T\mathbf{1} -\mathbf{W}^T\mathbf{X}\mathbf{1})$, where $n$ is number of samples in a dataset. Their proposed algorithm continuously updates $\mathbf{W}$, $\mathbf{b}$ and $\mathbf{Y}_U$ until convergence.

 Yamada et al. \cite{yamada2014high} proposed a non-linear FS method for high-dimensional datasets called Hilbert-Schmidt independence criterion least absolute shrinkage and selection operator (HSIC-Lasso), in which the most informative non-redundant features are selected using a set of kernel functions, where the solutions are found by solving a LASSO problem. The complexity of the original Hilbert-Schmidt FS (HSFS) is $O(n^4)$. In a recent work \cite{yamada2018ultra} called Least Angle Nonlinear Distributed (LAND), the authors have improved the computational power of the HSIC-Lasso. They have demonstrated via some experiments that   LAND and HSIC-Lasso  attain similar classification accuracies and dimension reduction. However, LAND has the advantage that it can be deployed on parallel distributed computing.  Another  kernel-based  feature selection method  is introduced in \cite{chen2017kernel}  using measures  of independence and minimizing the trace of the conditional covariance operator. It is motivated by selecting the features that maximally account for the dependence on the covariates' response.

In some recent  real-world applications, we need to deal with sequentially added dimensions in a feature space while the number of data instances is fixed. Yu et al. \cite{yu2016lofs} developed an open source Library of Online FS (LOFS) using state-of-the-art algorithms. The learning module of LOFS consists of two submodules, Learning Features added Individually (LFI) and Learning Grouped Features added sequentially (LGF).
The LFI module includes various FS methods including Alpha-investing \cite{zhou2006streamwise}, OSFS \cite{wu2010online}, Fast-OSFS \cite{wu2012online}, and SAOLA \cite{yu2014towards}  to learn features added individually over time, while the LGF module provides the group-SAOLA algorithm \cite{yu2016scalable} to  mine grouped features added sequentially. 

{In \cite{pierezan2018coyote}, the authors proposed a bio-inspired optimization algorithm called Coyote Optimization Algorithm (COA) simulating the behavior of coyotes. Using their new strategy, COA makes a balance between exploration and exploitation processes to solve continuous optimization problems. Very recently, the authors \cite{de2020binary} upgraded their method by proposing a binary version of COA called Binary Coyote Optimization Algorithm (BCOA), which is a wrapper feature selection method.

There is a great interest in the applications of FS methods in disease diagnoses and prognoses. For example, Parkinson's disease (PD) is  a critical neurological disorder and its diagnoses in its initial stages is extremely complex and time consuming. Recently,   voice recordings and handwritten drawings of PD patients are used to extract a subset of important features using FS  algorithms to  diagnose PD \cite{ali2019early, ashour2020novel, sharma2019diagnosis, kim2020deep,yaman2020automated} with a good success rate.

	\section{Proposed approach}\label{approach}
	
	Consider a dataset $D$ consisting of $m$ samples where each sample contains $n + 1$ features. Let us denote by $A$ the first $n$ columns of $D$ and by $\mathbf{b}$ the last column. We also denote by $\mathbf{F}_i$ the $i$-th feature (column) of $A$. We shall first consider eliminating the irrelevant features.  Throughout, by the norm of a vector, we always mean its 2-norm.
	Recall that 
	\begin{displaymath}
	||A||=\text{Sup}_{\mathbf{x}\neq 0} \frac{||A\mathbf{x}||}{||\mathbf{x}||}=\text{Sup}_{||\mathbf{x}||=1} ||A\mathbf{x}||
	\end{displaymath}

	Denote by $\sigma_1\geq \sigma_2\geq \cdots \geq  \sigma_r $ the singular values of $A$, where $r=\text{min}(m,n)$. The smallest non-zero singular value of $A$ is denoted by $\sigma_{\text{min}}$ and the greatest of the $\sigma_i$ is also denoted by $\sigma_{\text{max}}$. It is well-known that $||A||_2=\sigma_{\text{max}}$. Recall that $A$ admits a singular value decomposition (SVD) in the form $A=U\Sigma V^T$, where $U$ and $V$ are orthogonal matrices and  $\Sigma=\text{diag}(\sigma_1, \ldots, \sigma_r, 0, \ldots, 0 )$ is an $m\times n$ diagonal matrix. Here, $V^T$ denotes the transpose of $V$.

	We first normalize the columns of $A$ so that each $\mathbf{F}_i$ has norm 1. Then, we solve the linear system $A\mathbf{x} = \mathbf{b}$ by using the method of least squares (see theorem \ref{ls-solutions}). Here, $\mathbf{x}=[x_1 \cdots x_i \cdots x_n ]^T$.  The idea is to select a small number of columns of $A$ that can be used to approximate $\mathbf{b}$.  Since $A\mathbf{x} = \mathbf{b}$ may not have a unique solution, instead we consider a broader picture by solving the least squares problem $\text{min}_{\mathbf{x}} ||A\mathbf{x} - \mathbf{b}||_2$ whose solution is given in terms of pseudo-inverse   and  SVD of $A$. The following result is well-known, see \cite{ben2003generalized}.
	
	\begin{thm}[All Least Squares Solutions]\label{ls-solutions} 
		Let $A$ be an $m\times n$ matrix and $\mathbf{b}\in \mathbb{R}^m$. Then all the solutions of $\text{min}_{\mathbf{x}} ||A\mathbf{x} - \mathbf{b}||_2$ are of the form $\mathbf{y}=A^{+}\mathbf{b}+\mathbf{z}$, where $\mathbf{z}\in \ker(A)$. Furthermore, the unique solution whose 2-norm is the smallest is given by $\mathbf{x}=A^{+}\mathbf{b}$.
	\end{thm}

	In other words, we can approximate the label column $\mathbf{b}$ as a linear combination $x_1\mathbf{F}_1+\cdots +x_n\mathbf{F}_n$.  So each $|x_i|$ can be viewed as an assigned  weight to $\mathbf{F}_i$. Given that each $\mathbf{F}_i$ has norm 1, if  $|x_i|$ is small compared to others, then the vector $x_i\mathbf{F}_i$ will have a negligible effect on  $\mathbf{b}$. It then makes sense to filter out those features whose weights are very small. In other words, we shrink the weights of irrelevant features to zero. 
	
	In this paper, we shall mostly focus on datasets where $m<<n$.
	Of special interest to us are genomic datasets where there are usually  tens or hundreds of samples compared to  tens of thousands of genes. The matrix $A$ in these datasets has full row-rank because gene expressions of different samples are independent of each other. Since $\mathbf{b} \in \mathbb{R}^m$, it is enough to identify only $m$ independent columns of $A$. Intuitively, it makes sense to eliminate the columns that are less important.  
	
	We prove below how weight of each feature  $\mathbf{F}_i$ is directly  affected by the relevance of $\mathbf{F}_i$  to $\mathbf{b}$. 
	Suppose that  $A$ is an $m\times n$ of full row-rank and 
	consider the  SVD of $A$ as  $A=U\Sigma V^T$. Let 
	$U=\left[
	\begin{array}{c|c|c}
	\mathbf{u}_{1} & \cdots & \mathbf{u}_{m}
	\end{array}
	\right]
	$. 
	Note that $\mathbf{u}_{1} , \ldots,  \mathbf{u}_{m}$ form an orthonormal basis of $\mathbb{R}^m$.

	\begin{thm}\label{irrelevants}
		Let $A$ be a full row-rank matrix and denote by $\mathbf{x}=[x_1 \ldots x_n ]^T$ the least squares solutions to $A \mathbf{x}=\mathbf{b}$. Then, each component $x_i$ of $\mathbf{x}$ is   given by $x_i=\langle \mathbf{F}_i, \mathbf{z}\rangle$, where $\mathbf{z}=U[\langle \mathbf{u}_1,\mathbf{b}\rangle/\sigma_1^2\, \cdots\, \langle \mathbf{u}_m,\mathbf{b}\rangle/\sigma_m^2 ]^{T}$.
	\end{thm}
	\Proof Since $A$ is full row-rank,  the right inverse of $A$ is $A^{+}=A^T(AA^T)^{-1}$. Consider the SVD of $A$ as 
	$A=U\Sigma V^T$. Then $AA^T=U\Sigma \Sigma^T U^T=\sum_{i=1}^m \sigma_i^2 u_i u_i^T$.
	Note that the solution of $A\mathbf{x}=\mathbf{b}$ with the smallest norm is $\mathbf{x}=A^{+}\mathbf{b}=A^T(AA^T)^{-1}\mathbf{b}$.
	Let $\mathbf{z}=(AA^T)^{-1}\mathbf{b}$. So, $\mathbf{b}=AA^T \mathbf{z}=\sum_{i=1}^m \sigma_i^2 u_i u_i^T \mathbf{z}$. Since the $\mathbf{u}_i$s are orthonormal, we get
	\begin{align}\label{bk-zk}
	\langle \mathbf{u}_k,\mathbf{b}\rangle=\sigma_k^2\langle \mathbf{u}_k, \mathbf{z}\rangle, \quad k=1, \ldots, m.
	\end{align}
	Since $A$ has full row-rank, we have $\sigma_k>0$, for all $k$.
	Let $\bar b_1, \ldots, \bar b_m$ be the coordinates of $\mathbf{b}$ with respect to the  basis $\mathbf{u}_{1} , \ldots,  \mathbf{u}_{m}$  of $\mathbb{R}^m$.  
	Similarly, let $\bar z_1, \ldots, \bar z_m$ be the coordinates of $\mathbf{z}$ with respect to this  basis. So, $\mathbf{b}=U[\bar b_1\, \ldots\, \bar b_m]^T$ and  $\mathbf{z}=U[\bar z_1\, \ldots\, \bar z_m]^T$. Equation \eqref{bk-zk} can be written as $\bar b_k=\sigma_k^2 \bar u_k$, for each $1\leq k\leq m$. 
	On the other hand, $\mathbf{x}=A^{+}\mathbf{b}=A^T(AA^T)^{-1}\mathbf{b}=A^T \mathbf{z}$. 
	Since $\mathbf{z}=U[\bar z_1\, \ldots\, \bar z_m]^T$, we deduce that 
	$x_i= \langle \mathbf{F}_i, \mathbf{z}\rangle$, for each $i$. \qed

	We note that   the extent to which  a feature is relevant to $\mathbf{b}$ also depends on how important the other features are in determining $\mathbf{b}$. This fact is reflected in Theorem \ref{irrelevants} by taking into account the singular values of $A$ that encode  part of the information about $A$.	Also, the definition of relevancy is not quantitative and one has to set a threshold for the degree of relevancy.
	We set a dynamic threshold by  calculating  the average of all local maxima in $\mathbf{x}$ and remove those features that their corresponding value $|x_i|$ is smaller than the threshold.  
	In a sense, the threshold is set so that rank of the reduced matrix is still the same as of the original $A$. So, in the reduced matrix, we only keep features that have a higher impact on $\mathbf{b}$ and yet the reduced matrix retains the same prediction power as $A$ in approximating $\mathbf{b}$.

	If $A$ is full row-rank then, it follows from Theorem \ref{ls-solutions}  that the solution $\mathbf{x}^R$ of smallest 2-norm to  the system  $A\mathbf{x}=\mathbf{b}$  is in the row space of $A$. 
	So, there is a vector $\mathbf{y}$ such that $\mathbf{x}^{R}=-A^T\mathbf{y}$. Hence, $\mathbf{x}^{R}$ satisfies $\mathbf{x}^R+A^T\mathbf{y}=0,	A\mathbf{x}^R=\mathbf{b}$.
	In other words, $\mathbf{x}^R$ is part of the solution to the non-singular linear system 
	$$ \left(
	\begin{array}{ c c }
	I & A^T \\
	A & 0
	\end{array} \right)
	\left(
	\begin{array}{ c c }
	x^R \\
	y
	\end{array} \right)
	=\left(
	\begin{array}{ c c }
	0 \\
	b
	\end{array} \right)
	$$

		 Next, we  show how we can detect correlations between features. 
	Recall that a perturbation of $A$ is of the form $A+E$, where $E$ is a random matrix with the normal distribution.  
	We choose $E$ to be a random matrix such that $||E||_2\approx 10^{-s}\sigma_{\text{min}}(A)$, for some $s\geq 0$. We set $s=3$ where our estimates are correct up to a magnitude of $10^{-3}$.

		\begin{thm}\label{perturb} Let $\mathbf{x}$ and $\mathbf{{\tilde x}}$ be solutions of $A\mathbf{x}=\mathbf{b}$ and $(A+E)\mathbf{{\tilde x}}=\mathbf{b}$, where $E$ is a  perturbation such that such that $||E||_2= 10^{-s}\sigma_{\text{min}}(A)$. 
		If a feature $\mathbf{F}_i$ is independent of the rest of the features, then $|x_i-\tilde x_i|\approx0$. Furthermore,  suppose that  features $\mathbf{F}_j$ and $\mathbf{F}_k$ correlate, say $\F_j=c\F_k$ for some scalar $c$. If $\mathbf{F}_j$ and $\mathbf{F}_k$ are independent from the rest of the features, then $c=\frac{x_k-\tilde x_k}{x_j-\tilde x_j}$.
	\end{thm}
	\Proof
	From $A\mathbf{x}=\mathbf{b}$ and $(A+E)\mathbf{{\tilde x}}=\mathbf{b}$, we get $A(\mathbf{x}-\mathbf{{\tilde x}})=E\mathbf{{\tilde x}}$. 
	Consider the SVD of $A+E$ which is of the form $A+E=U\Sigma V^T$. So, 
	$\mathbf{{\tilde x}}=V\Sigma^{-1}U^T \mathbf{b}$. Since $U$ and $V$ are orthogonal and for orthogonal matrices we have $|| U\mathbf{v}||_2=||\mathbf{v}||_2$, we get
	\begin{align*}
	||\mathbf{{\tilde x}}||_2=||V\Sigma^{-1}U^T \mathbf{b}||_2&=||\Sigma^{-1} \mathbf{b}||_2\\
	&\leq ||\Sigma^{-1}||_2||\mathbf{b}||_2=\frac{1}{\sigma_{\text{min}}(A+E)}\\
	&\leq \frac{1}{-||E||_2+\sigma_{\text{min}}(A)}.
	\end{align*}
	Hence,
	$||E\mathbf{{\tilde x}}||_2\leq ||E|| || \mathbf{{\tilde x}}||_2\leq  10^{-s}$
	and we deduce that 
	\begin{align*}
	(x_1{-}\tilde x_1)\mathbf{F}_1+\cdots +(x_t{-}\tilde x_t)\mathbf{F}_t+\cdots +(x_n{-}\tilde x_n)\mathbf{F}_n\approx 0.
	\end{align*}
	
	Now, if a feature, say $\mathbf{F}_i$, is independent of the rest of features, then it follows  that $|x_i-\tilde x_i|\approx0$. Furthermore, since $\mathbf{F}_j$ and $\mathbf{F}_k$ are independent from the rest of the features, we must have 
	\begin{align*}
	(x_j-\tilde x_j)\mathbf{F}_j+(x_k-\tilde x_k)\mathbf{F}_k\approx 0.\label{depn-rel}
	\end{align*}
	So, $\mathbf{F}_j=\frac{x_k-\tilde x_k}{x_j-\tilde x_j}\F_k$. Hence, $c=\frac{x_k-\tilde x_k}{x_j-\tilde x_j}$, as required.
		\qed
	
	Theorem \ref{perturb} shows how we can filter out irrelevant features by looking at the components of  $\mathbf{x}-\mathbf{\tilde x}$ that are close to zero. Also, as we mentioned before, we normalize the columns of $A$ so that each $\mathbf{F}_i$ has norm 1. 
	So, if in the raw dataset $\F_j$ and $\F_k$ correlate then after normalization we must have 
	$\F_j'=\pm \F_k'$. Here, $\F_j'=\frac{\F_j}{|| \F_j||}$. So, by 	Theorem \ref{perturb}, if 
	$\F_j$ and $\F_k$ are independent from other features, we must have  $|x_k-\tilde x_k|=|x_j-\tilde x_j|$. We explain these notions further in a synthetic dataset.  
	Consider  a synthetic dataset with 22 features and 100 samples and the label column  $\mathbf{b}$ which we set  as $\mathbf{b}=3\mathbf{F}_{19}+5\mathbf{F}_{17}+2\mathbf{F}_{20}$. The first 20 features of this dataset are generated randomly in the interval of -1 and 1. The correlations between remaining features are set as follows: 
	$\mathbf{F}_{21}= 2\mathbf{F}_{18}+4\mathbf{F}_{19}$ and $\mathbf{F}_{22}=3\mathbf{F}_{20}$. 
	First,  we normalize $A$. Then    solve $A\mathbf{x}=\mathbf{b}$ and   $(A+E)\mathbf{\tilde x}=\mathbf{b}$ and calculate 
	$\Delta\mathbf{x}$ as shown in Table \ref{synthetic}.

	\begin{table}[H]
				\centering
		\caption{Perturbation of the synthetic Dataset}	\label{synthetic}
				\begin{tabular}{l c c c c c c}
					\hline
					\bf{} &\bf{x} & $\tilde{\mathbf{x}}$& $\Delta\mathbf{x}=|\mathbf{x}-\mathbf{\tilde x}|$\\\hline
					$\F_{1}$\dots $\F_{16}$  & $\leq$3.0987e-14 &$\leq$2.6907e-05 &$\leq$4.7316e-05 \\
					$\F_{17}$ & 29.1715& 	  29.1715 & 2.7239e-05\\
					$\F_{18}$ & -3.4494& 	-10.2466 &6.7972\\
					$\F_{19}$ & 9.9339&  -3.1806 & 13.1145\\
					$\F_{20}$ & -5.3307&  -6.0073& 0.6766\\
					$\F_{21}$ & 7.3630&   21.8723& 14.5093\\
					$\F_{22}$ & -5.3307&   -4.6541 &0.6766\\
					\hline
				\end{tabular}
	\end{table}
	Let $\Delta \mathbf{x}=|\mathbf{x}-\tilde{\mathbf{x}}|$ and denote its $i$-th component with $\Delta x_i$. Since  each of  $\F_{1}, \ldots,  \F_{17}$ are independent 
	from the other features, as we expected, we have $\Delta x_i\approx 0$, for all $1\leq i\leq 17$. However, $\F_{17}$ is relevant because it correlates with $\mathbf{b}$. 
	Indeed, $\F_{17}$ is a very important feature because we cannot make up for its loss using other features. Hence, we should be able to distinguish and preserve $\F_{17}$ from other $\F_i$ for which $\Delta x_i=0$. This can be accomplished by noting  that 
	 irrelevant features have smaller $|x_i|$ compared to other features as can be seen from Table \ref{synthetic}. 
	 
	  Since $\mathbf{F}_{22}=3\mathbf{F}_{20}$, these features correlate and are independent from other features. By  normalization, we get  $\mathbf{F}_{22}'=\mathbf{F}_{20}'$. So,  we expect to have  $\Delta x_{20}\approx \Delta x_{22}$ as shown in Table \ref{synthetic}. We deduce that 
	   $\mathbf{F}_{22}$ and $\mathbf{F}_{20}$ are dependent and so we should only choose one of them.

	 Finally, after normalization and rewriting  the relation  $\mathbf{F}_{21}= 2\mathbf{F}_{18}+4\mathbf{F}_{19}$, we obtain  $\F_{21}'\approx \F_{19}'$ modulo $\F_{18}'$. 
	The reason for  this is because norms of  $\F_{21}'$ and  $\F_{19}'$ outweigh norm of  $\F_{18}'$. This is confirmed in Tabel  \ref{synthetic} as  $\Delta x_{19}$ and  $\Delta x_{21}$ are closest to each other compared  to the others. So, $\mathbf{F}_{19}$ and  $\F_{21}$  fall into the same cluster of $\Delta \mathbf{x}$. This means that one of $\mathbf{F}_{19}$ or   $\F_{21}$ must be removed as a redundant feature. We  shall now explain the clustering process of $\Delta \mathbf{x}$.

	By Theorem \ref{perturb}, if features $\mathbf{F}_i$ and $\mathbf{F}_j$ correlate, then 
	the differences $\Delta x_i$ and $\Delta x_j$ are almost the same. That is, the correlations between features are encoded in 
	$\Delta \mathbf{x}$.
	Now, we sort $\Delta \mathbf{x}$ and  obtain  a stepwise function where each step can be viewed as a cluster consisting of features that possibly correlate with each other. To find an optimal number of steps, it makes sense to smooth $ \Delta \mathbf{x}$  where we view   $\Delta \mathbf{x}$   as a signal and use a simplified least-squares method called Savitsky-Golay  smoothing filter \cite{schafer2011savitzky}. Figure \ref{Smooth} exhibits how the smoothening process on   $\Delta \mathbf{x}$ preserves its whole structure without changing the trend. 
	
	We note that the converse of Theorem \ref{perturb} may not be true in general. That is $\Delta x_i$ and $\Delta x_j$ being the same does not necessarily imply that $\mathbf{F}_i$ and $\mathbf{F}_j$ correlate. Hence, in the next step, we want to further break up each cluster of  $\Delta \mathbf{x}$ into sub-clusters.  There are several ways to accomplish this step and one of the most natural ways is to use entropy of features.
	
	Generally, entropy is a key measure for information gain and it is capable of quantifying the disorder or uncertainty of random variables. Also, entropy effectively scales the amount of information that is carried by random variables. Entropy of a feature $\mathbf{F}$ is defined as follows:
	\begin{equation}\label{ent}
	H(\mathbf{F})= -\sum_{k=1}^{m}{f_k\log f_k}
	\end{equation}
	where $m$ is the number of samples and $f_k$  is the frequency with which $\mathbf{F}$ assumes the $k$-th value in the observations.

	\begin{figure}
		\centering
		\includegraphics[height=4cm,width=.9\textwidth]{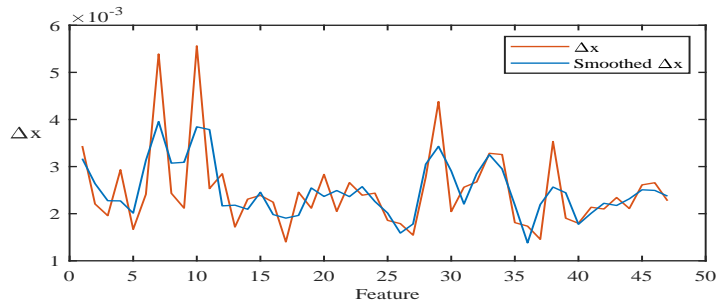}
		\caption{$\Delta \mathbf{x}$ vs. smoothed $\Delta \mathbf{x}$}
		\label{Smooth}
	\end{figure}

	Figure \ref{steps}(a) shows  clustering the set of all features based on    $\Delta \mathbf{x}$, and then     a typical cluster   splits into sub-clusters using entropy as shown in Figure \ref{steps}(b). 
	To do so, we sort the features of a cluster based on their entropy which yields another step-wise function. At this stage, we pick one candidate feature from each sub-cluster based on the corresponding values $|x_i|$.
	Finally, the selected features are ranked based on both their entropies and the corresponding  $|x_i|$'s.
	The final sorting of the features is an amalgamation of these two rankings.

	\begin{figure}
		\centering
		\includegraphics[height=7cm,width=.9\textwidth]{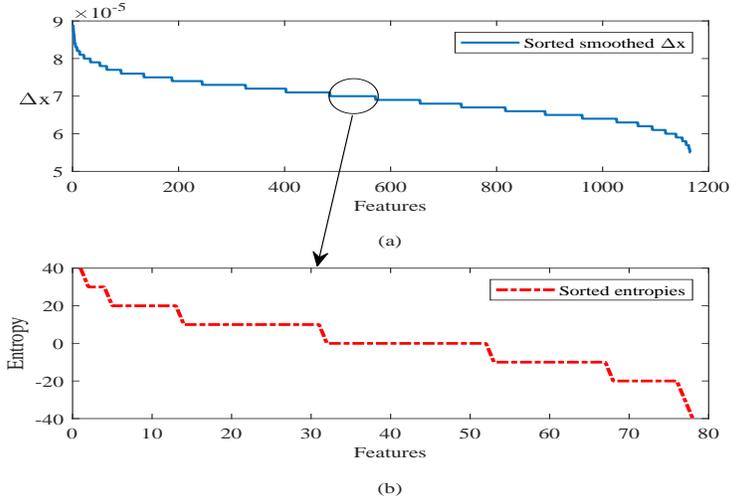}
		\caption{(a) Sorted smoothed $\Delta \mathbf{x}$ (b) Sorted entropies of the magnified cluster}
		\label{steps}
	\end{figure}

\subsection{Noise-robustness and stability}

In real datasets, it is likely that $D$ involves some noise.  For example, in genomics,  it is conceivable that through the process of preparing a genomic dataset,  some error/noise  is included and as such  the dataset $D$ is noisy. We note that the label column $\mathbf{b}$ is already known to us (and without noise). So instead of $D=[A\mid \mathbf{b}]$ we deal with $D=[A_1\mid \mathbf{b}]$, where $A_1=A+E_1$ and $|| E_1||_2$ is small ($|| E_1||_2=10^{-s}\sigma_{\text{min}}(A)$). A perturbation of $A_1$ is of the form $\tilde{A}_1=A_1+E_2$, where $|| E_2||_2=10^{-s}\sigma_{\text{min}}(A)$.  Our aim is to show that if certain columns of $A$ correlate,  then so do the same columns of $A_1$  and vice versa. 

\begin{thm}\label{noise}
	Let $ \tilde{\mathbf{x}}, {\mathbf{\tilde y}}$ be solutions of  $A_1\tilde{\mathbf{x}}=\mathbf{b}$
	and $\tilde A_1{\mathbf{\tilde y}} =\mathbf{b}$, respectively.
	Suppose that $S'=\{\F_1, \ldots, \F_t\}$ is set of  columns of $A$  such that  $\sum_{i=1}^t c_i \F_i=0$, for some non-zero $c_i$. If 
	\begin{enumerate}
		\item  any subset of $S'$ is linearly independent,
		\item $\F_1, \ldots, \F_t$  are linearly independent from the remaining columns of $A$.
	\end{enumerate}
	Then the vectors 
	$
	\begin{bmatrix}
	c_1 & \cdots &	c_t
	\end{bmatrix}
	$
	and 
	$
	\begin{bmatrix}
	\tilde x_1-\tilde y_1 & \cdots &
	\tilde x_t-\tilde y_t
	\end{bmatrix}
	$
	are proportional.
\end{thm}
\Proof
From $(A+E_1)\tilde{\mathbf{x}}=\mathbf{b}$ and $(A+E_1+E_2)\tilde{\mathbf{y}}=\mathbf{b}$, we get 
$A(\tilde{\mathbf{x}}-\tilde{\mathbf{y}})=-E_1\tilde{\mathbf{x}}_1+(E_1+E_2)\tilde{\mathbf{y}}$. 
Similar arguments as in the proof of Theorem \ref{perturb} can be used to show that 
\begin{align*}
||-E_1\tilde{\mathbf{x}}+(E_1+E_2)\tilde{\mathbf{y}}||&\leq ||-E_1\tilde{\mathbf{x}}_1||+ ||(E_1+E_2)\tilde{\mathbf{y}}|| \\
&\leq \frac{1}{10^{s}-1}+\frac{2\cdot 10^{-s}}{-2\cdot 10^{-s}+1}\\
&\leq \frac{1}{10^{s}-1}+\frac{2}{ 10^{s}-2}\\
\approx 3\cdot 10^{-s}
\end{align*}

We deduce  that 
\begin{align}\label{linear-com}
(\tilde x_1-\tilde y_1)\F_1+\cdots +(\tilde x_t-\tilde y_t)\F_t+\cdots +(\tilde x_n-\tilde y_n)\F_n\approx 0.
\end{align}

Since $\F_1, \ldots,\F_t$  are linearly independent from the rest of features in $S$,   we get 
\begin{align}
(\tilde x_1-\tilde y_1)\F_1+\cdots +(\tilde x_t-\tilde y_t)\F_t\approx 0.\label{depn-rel}
\end{align}
Now, if 
$
\begin{bmatrix}
c_1 & \cdots &	c_t
\end{bmatrix}
$
and 
$
\begin{bmatrix}
\tilde x_1-\tilde y_1 & \cdots &
\tilde x_t-\tilde y_t
\end{bmatrix}
$ 
are not proportional, we can use Equation \eqref{depn-rel} and our first hypothesis to get a dependence relation of a shorter length between the elements of $S'$, which would contradict our assumption that any proper subset of $S'$ is linearly independent. The proof is complete. \qed

	We  also remark that our method is insensitive to shuffling of the dataset $D$. That is,  if we exchange rows (or columns), there is  an insignificant change in $\Delta \mathbf{x}$. We have demonstrated this fact through experiments in Tables \ref{Standard_deviation}; we offer a proof as follows.
	
	\begin{thm}\label{shuffle} 
		DRPT is insensitive to permuting rows or columns.
	\end{thm}
	\Proof
		We show this for permutation of rows  and a similar argument  can be made for permuting columns. Suppose that $D_1=[A_1\mid  \mathbf{b}_1]$ is obtained from $D$ by shuffling rows. First, assume that only two rows are exchanged. Then there exists an elementary matrix $T$ such that $TA=A_1$ and $T \mathbf{b}= \mathbf{b}_1$.
		Since $T$ is invertible, it follows that $A\mathbf{x}= \mathbf{b}$ if and only if $A_1\mathbf{x}= \mathbf{b}_1$. For the general case, we note that every shuffling is a composition  of elementary matrices.

	\subsection{Algorithm}

	The Flowchart and algorithm   of DRPT are as follows. 
	The MATLAB\textsuperscript \textregistered implementation of DRPT is publicly available in GitHub \footnote{http://github.com/majid1292/DRPT}.

	\savebox\DontUseBreqn{$\mathbf{x}$}
	\savebox\DontUseBreqn{$\tilde{\usebox\DontUseBreqn}$}
	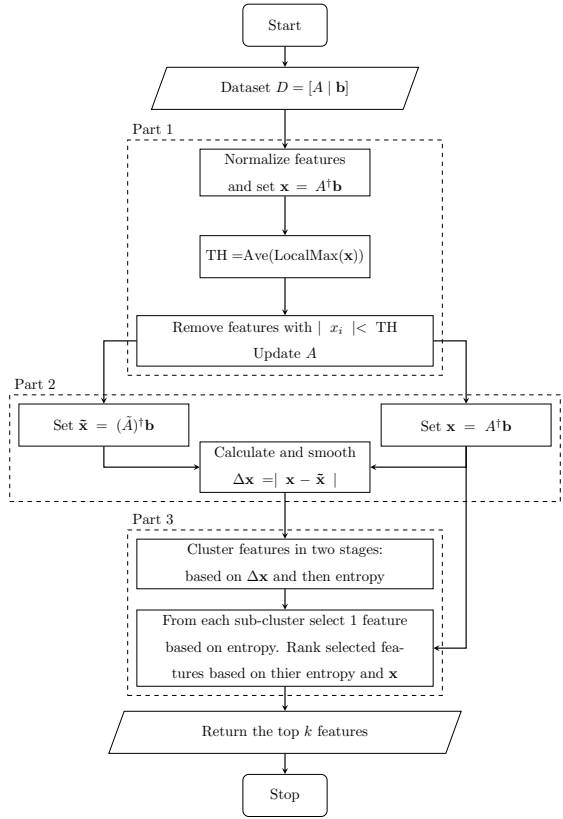
\begin{figure}[H]\label{flowchart}
		\centering
		\scalebox{.56}{
			\begin{tikzpicture}[node distance=1.5cm]
			\tikzstyle{block} = [%
			rectangle, draw,thick,fill=blue!0,
			text centered, minimum height=1em,
			execute at begin node={\begin{varwidth}{40em}},
			execute at end node={\end{varwidth}}]

			\node (start) [startstop] {Start};
			\node (in1) [io, below of=start] {Dataset $D=[A\mid  \mathbf{b}]$};
			\node (pro1) [process, below of=in1, yshift=-.5cm] {Normalize features  and set   $\mathbf{x}=A^{\dagger}\mathbf{b}$};
			\node (pro2) [process, left of=pro1, yshift=-6cm, xshift=-2.8cm] {Set  $\mathbf{\tilde x}=(\tilde A)^{\dagger}\mathbf{b}$}; 
			\node (pro3_1) [process, below of=pro1, yshift=-.5cm]   {					$\text{TH}=$Ave(LocalMax($\textbf{x})$)\\
			};		
			\node (pro3) [process1, below of=pro3_1, yshift=-.5cm]   {	
				Remove features with $\mid x_i\mid <\text{TH}$
				\\ Update $A$};

			\node (pro4) [process, right of=pro1, yshift=-6cm, xshift=2.8cm] { Set  $\mathbf{x}=A^{\dagger}\mathbf{b}$};
			
			\node (pro5) [process, below of=pro3, yshift=-1.5cm] {Calculate and smooth 
				$\Delta \mathbf{x}= \mid\mathbf{x}-\mathbf{\tilde x}\mid $};
			\node (pro6) [process1, below of=pro5, yshift=-.8cm] {Cluster features in two stages: based on  
				$\Delta \mathbf{x}$ and   then  entropy};
			\node (pro7) [process1, below of=pro6, yshift=-.5cm] {From each sub-cluster select 1 feature based on entropy. Rank selected features based on thier entropy and $\mathbf{x}$};
			\node (out1) [io, below of=pro7, yshift=-.5cm] {Return the top $k$ features};
			\node (stop) [startstop, below of=out1] {Stop};
			
			\node[container1, fit=(pro1) (pro3)] (or) {};
			\node at (or.north west) [above right,node distance=0 and 0] {Part 1};
			
			\node[container1, fit=(pro2) (pro4) (pro5)] (or) {};
			\node at (or.north west) [above right,node distance=0 and 0] {Part 2};
			
			\node[container1, fit=(pro6) (pro7)] (or) {};
			\node at (or.north west) [above right,node distance=0 and 0] {Part 3};
			
			\draw [arrow] (start) -- (in1);
			\draw [arrow] (in1) -- (pro1);
		    \draw [arrow] (pro3) -| (pro2);
			\draw [arrow] (pro1) -- (pro3_1);
			\draw [arrow] (pro3_1) -- (pro3);
			\draw [arrow] (pro3) -| (pro4);
			\draw [arrow] (pro2) |- (pro5);
			\draw [arrow] (pro4) |- (pro7);
			\draw [arrow] (pro4) |- (pro5);
			\draw [arrow] (pro5) -- (pro6);
			\draw [arrow] (pro6) -- (pro7);
			\draw [arrow] (pro7) -- (out1);
			\draw [arrow] (out1) -- (stop);
			\end{tikzpicture}
		}
		\caption{Flowchart of Dimension Reduction based on Perturbation Theory (DRPT)}
	\end{figure}
	
	\begin{algorithm}[H]
		\label{method}
		\caption{Dimension reduction based on perturbation theory (DRPT)}
		\begin{algorithmic}
			\State \textbf{Input:} $D=[A\mid \textbf{b}], k$
			\State \textbf{Output:} A subset of features of size $k$
			
			\State \Comment { \textit{//***Part1: Irrelevant Feature Removal***                    }}
			\State Normalize columns of $A$ within $[0, 1]$
			\State $\textbf{x}=A^{+} \textbf{b}$
			\State $\text{TH}=$Average(LocalMaxima of $\textbf{x}$)
			\State $I=\emptyset$
			\For{each $x_i \in$ \textbf{x}}
			\If{$ x_i \geq \text{TH} $}
			\State $I = I \cup {i}$
			\EndIf
			\EndFor
			\State $A \leftarrow A[I]$
			\State \Comment {\textit{//***Part2: Detecting Correlations***                      } }
			\State $s = 3$
			\State $(m,n)=\text{Size}(A)$
			\State $minSVD = Min$(singular value of $A$)
		
			\State $\textbf{x}=A^{+} \textbf{b}$
			\State	$t = 10^{-s} \cdot minSVD$
			\State	Set $E$ be a random $m\times n$ matrix with uniform dist.  in the interval (0,1)

			\State 	$E = t .* E$
			\State	$\tilde{A} = A + E$	
			\State	$\Delta \textbf{x} = |(\tilde{A})^{+}  \textbf{b} - \textbf{x}|$
		
			\State $\Delta \textbf{x}= \text{Smooth}(\Delta \textbf{x})$
			\State \Comment {\textit{//***Part3: Ranking  Features***                            }}
			\While{ $z \in \text{unique}( \Delta \textbf{x})$ }
			\State $ \text{CL}=\{ \textbf{F}_i\mid  |x_i-\tilde{x}_i|=z$\}

			\For { $h\in \text{unique} (H(\text{CL}))$}
			\State $ \text{subCL}=\{ \textbf{F}_i\in  \text{CL} \mid  H(\textbf{F}_i)=h\}$
			\State $\text{Pick } \textbf{F}_i \text{ in subCL with } |x_i|=  \max\textbf{x}_{\text{subCL}}$
			\State $\text{Output} \leftarrow \text{Output} \cup  \textbf{F}_i$ 
			\EndFor
			\EndWhile
			\State $Ranked$ $Output$ $\leftarrow$ Rank ($Output$, \{$H(F)$  \&  $\textbf{x}$\})
			\State \textbf{Return:} the top $k$ features
		\end{algorithmic}
	\end{algorithm}

	\subsection{Complexity} The complexity of our proposed method is dominated by the complexity of the SVD which is $O(mn^2, m^2n)$, since  the inverse of perturbed $\tilde{A}$ is calculated using SVD.
	
	\section{Experimental Results}\label{Results}
	We compared our method with seven state-of-the-art FS methods, namely minimal-redundancy-maximal-relevance criterion (mRMR) \cite{peng2005feature}, least angle regression (LARS) \cite{efron2004least}, Hilbert-Schmidt Independence Criterion Lasso (HSIC-Lasso) \cite{yamada2014high}, Fast Online Streaming FS (Fast-OSFS) and Scalable, Accurate Online FS (group-SAOLA) \cite{yu2016lofs}, Conditional Covariance Minimization (CCM) \cite{chen2017kernel} and Binary Coyote Optimization Algorithm (BCOA) \cite{de2020binary} . We used MATLAB \textsuperscript\textregistered  implementations of LARS and LASSO by Sj{\"{o}}strand \cite{Sjostrand}, HSIC-Lasso and BCOA by their authors, Fast-OSFS and  group-SAOLA given in the open source library \cite{yu2016lofs}. The CCM method is implemented in Python by their authors and its code available at  GitHub\footnote {https://github.com/Jianbo-Lab/CCM}.

	To have a fair comparison among different FS methods, we read the datasets by the same function and use a stratified partitioning of the dataset so that  70\%   of each class is selected for FS. Then we use SVM and RF classifiers implemented in MATLAB\textsuperscript \textregistered, to evaluate the selected subsets of features on the remaining 30\% of the dataset. We have used linear kernel in  SVM (default setting of SVM  in MATLAB\textsuperscript \textregistered) and as for RF, we  set 30 as the number of trees and the other parameters have default values.

	\subsection{Datasets}\label{Datasets}
	We select a variety of dataset from Gene Expression Omnibus (GEO) \footnote{https://www.ncbi.nlm.nih.gov/geo/}  and dbGaP \footnote{https://www.ncbi.nlm.nih.gov/gap/}  to perform FS and classification. The specifications of all datasets are given in Table \ref{data}. 
	
	\begin{table}[!ht]
		\caption{Dataset Specifications}
		\centering \label{data}
		\begin{adjustbox}{width=\columnwidth,center}
			\resizebox{\textwidth}{!}{
				\begin{tabular}{l c c c c c| c| c| c}
					\hline
						\multirow{2}{*}{\bf{Dataset}} &	\multirow{2}{*}{\bf{Samples}} &\multirow{2}{*}{\bf{\# Original F}}&	\multirow{2}{*}{\bf{\# Cleaned F}}&\multirow{2}{*}{\bf{\# Labels}}&\multicolumn{4}{c}{\textbf{Proportion of labels}} \\
					&  & & &&\textbf{1}&\textbf{2}&\textbf{3}&\textbf{4} \\\hline
					GDS1615 & 127  & 		22,282&          		13,649 & 3& 33\%& 20.5\%& 46.5\%&--\\
					GDS3268 & 202 & 		44,290& 				29,916& 2&36.1\%  &63.9\% &-- &--\\
					GDS968 & 171 &		 12,625&			9,117 & 4& 26.3\% &26.3\% &22.8\% & 24.6\%\\
					GDS531 & 173 & 		12,625&				9,392& 2& 20.8\% &79.2\% &-- &--\\
					GDS2545 & 171 &		 12,625&				9,391 &4& 10.6\% & 36.8\%&38\% &14.6\%\\
					GDS1962 & 180 &		 54,675&				29,185 &4& 12.8\% & 14.4\%& 45&27.8\%\\
					GDS3929 & 183 &		 24,526& 			19,334 & 2&69.9\%  &30.1\% &-- &--\\
					GDS2546 & 167 &		 12,620& 			9,583&4& 10.2\% &35.3\% & 39.5\%&15\%\\
					GDS2547 & 164 & 		12,646&	 	9,370&4&10.4\%  &35.4\% &39\% &15.2\%\\
					NeuroX & 		11,402 		& 535,202 & 			267,601&2&48.6\%  &51.4\% &-- &--\\
					\hline
				\end{tabular}
			}
		\end{adjustbox}
	\end{table}

	All the GEO datasets are publicly available. 
	To pre-process the data, we develop an R code to clean and convert any NCBI dataset to CSV format  \footnote{http://github.com/jracp/NCBIdataPrep}.
	We use GEO2R \cite {GEO2R}   to retrieve the mappings between prob IDs and gene samples. Probe IDs without a gene mapping were removed. Expression values of each gene are  the average of expression values of  all mapped prob IDs  to that gene. We also  handle missing values with k-Nearest Neighbors (kNN) imputation method.

	The dataset NeuroX holds SNP information about subjects' Parkinson disease status and sociodemographic (e.g., onset age/gender) data. Parkinson's disease status coded as 0 (control) and 1 (case), from clinic visit using modified UK Brain Bank Criteria for diagnosis. The original NeuroX has 11402 samples, and it is only accessible by authorized access via dbGaP. It has 535202 features that each two sequence features are considered as a SNP. So after cleaning and merging features of  NeuroX, we use two subsets of 100 and 200 samples with 267601 SNPs (NX100 and NX200)  for this paper.

	\subsection{Parameters}
	
	A FS method that selects most  relevant and non-redundant features (Minimum Redundancy and Maximum Relevancy) is favorable in the sense that the top $k$ selected features retain most of the information about the dataset.  On the other hand, the top selected genes in a  genomic dataset
	must be further analyzed in wet labs to confirm the biological relevance of the genes to the disease. 
	For example,  authors  in \cite{zhou2005ls} first identified 50 top genes of a   Colon cancer dataset using their FS method. Then, they  selected the first 15 genes, because adding more genes would not result in significant changes to the prediction accuracy. Similar studies    \cite{gutkin2009slimpls, liu2018feature} suggest considering the top 50 features.  So, in Table \ref{results_features_accuracy}, we set $k=50$ to select a subset of  50 features using FS algorithms. Then,  for $t=1$ to $k$, we feed the first $t$ features to the classifier to find an optimal $t$ so that the subset of first $t$  features yields the highest accuracy.  This set up is applied across all FS methods. In Figure \ref{genomicresults},  we expand this idea by considering up to 90 features using each FS method. We can see that  there is small, incremental changes in classification accuracies when we increase the number of features from 50 to 90.  
	
	 We report the average classification accuracies  and average number of selected features over 10 independent  runs where the dataset is row shuffled in each run. We note the top $k$ selected features using a FS algorithm might differ over different runs because the dataset is row shuffled and so the training set changes on every run. Also,     optimal subset size for  SVM and RF might be different, in other words SVM might attain the maximum accuracy using the first 20 features while RF might attain its maximum using the first 25 features.

Both Fast-OSFS and group-SAOLA  have a parameter $\alpha$, which is a threshold on the significance level. The authors of the LOFS \cite{yu2016lofs} in their Matlab user manual \footnote{https://github.com/kuiy/LOFS/tree/master/LOFS_Matlab/manual} recommend  setting $\alpha=0.05$ or $\alpha=0.01$. However, our experiments based on these parameters showed   inferior  classification accuracies compared to other methods across all datasets given in Table \ref{data}. We also note that 
the running times of both  Fast-OSFS and group-SAOLA increased as we increased $\alpha$.

Experimenting with various values of $\alpha$, we realized that   increasing $\alpha$ from 0.01 to 0.5 exhibited clear improvement in classification accuracies on all datasets except NeuroX.  So, we set $\alpha=0.5$  for all datasets except NeuroX.

We also experienced that both Fast-OSFS and group-SAOLA on NeuroX may not    execute all the time when   $\alpha> 0.0005$; often errors were generated as part of a statistics test function. 
So,  for NeuroX, we set $\alpha=10^{-5}$ for both Fast-OSFS and group-SAOLA.

The group-SAOLA model has an extra parameter for setting the number of selected groups, \textit{selectGroups}. As there was no default or recommended value for this parameter, we obtained results by varying \textit{selectGroups} from 2 to 10  for all the datasets, and we chose the highest accuracy for each dataset. 
	
	\subsection{Hardware and Software} Our proposed method DRPT and other FS  methods in section \ref{Results} have been run on an IBM\textsuperscript \textregistered LSF 10.1.0.6 machine (Suite Edition: IBM Spectrum LSF Suite for HPC 10.2.0) with requested 8 nodes, 24 GB of RAM, and 10 GB swap memory using MATLAB\textsuperscript \textregistered R2017a (9.2.0.556344) 64-bit(glnxa64). Since  CCM  is implemented in  Python  and uses TensorFlow\cite{abadi2016tensorflow}, we requested 8 nodes, 120 GB of RAM, and 40 GB swap memory on the LFS machine using Python 3.6.}

	\subsection{Results.}\label{result}
	The average number of selected features and  average  classification accuracies over 10 independent runs using  SVM and RF on the datasets described in Section \ref{Datasets} are shown in Table \ref{results_features_accuracy}.  
	
	\begin{table}[!ht]
		\centering
		\caption{Superscript is average number of selected features and subscript is resulting classification accuracies (CA) based on SVM and RF using mRMR, LARS, HSIC-Lasso, Fast-OSFS,  group-SAOLA, CCM, BCOA and DRPT over 10  runs.}
		\centering
		\begin{adjustbox}{width=.92\columnwidth,center}
			\resizebox{\textwidth}{!}{
				\begin{tabular}{l c c c c c c c c c }
					\hline
					\multirow{2}{*}{\bf{\rotatebox{45}{\tiny Classifier}}} &\multirow{2}{*}{\bf{Dataset}} & \multicolumn{4}{c}{$(\text{\# of selected features})_{\text{classification accuracy}}$}\\
					&&\bf{mRMR}&\bf{LARS} &\bf{HSIC-Lasso} &\bf{Fast-OSFS} &\bf{group-SAOLA} &\bf{CCM}&\bf{BCOA}&\bf{DRPT} \\
					\hline
					\multirow{12}{*}{\bf{\rotatebox{45}{SVM}}}
					&GDS1615 &$(40.20)_{87.37}$&$(26.60)_{91.67}$&$(18.70)_{91.35}$&$(17.20)_{84.31}$&$(12.40)_{83.13}$	&$(29.20)_{80.82}$&$(33.90)_{84.90}$&$(37.00)_{\textbf{91.89}}$\\
					&GDS3268 &$(38.50)_{85.69}$&$(43.50)_{89.62}$& -- & $(35.90)_{87.89}$	&$(16.60)_{84.13}$	&$(38.65)_{85.82}$ &$(34.50)_{73.55}$&$(33.90)_{\textbf{90.45}}$\\
					&GDS968 &$(39.80)_{80.87}$&$(38.70)_{\textbf{83.73}}$& -- &$  (19.80)_{72.41}$	&$(14.10)_{70.53}$	&$(34.14)_{78.82}$&$(32.86)_{ 76.19}$&$(38.30)_{81.06}$\\
					&GDS531 & $(30.90)_{69.78}$&$(27.60)_{79.96}$&$(4.00)_{67.93}$&$(26.50)_{77.43}$	&$(11.60)_{77.70}$	&$(30.45)_{\textbf{80.82}}$&$(32.67)_{74.17}$&$(25.20)_{77.16}$\\
					&GDS2545 & $(34.00)_{75.90}$&$(33.70)_{79.02}$&$(33.8)_{76.40}$&$(18.80)_{74.95}$	&$(12.30)_{75.55}$	&$(30.11)_{70.82}$&$(29.85 )_{75.40}$&$(31.30)_{\textbf{83.23}}$\\
					&GDS1962 &$(39.50)_{65.12}$&$(32.50)_{76.56}$&$(31.5)_{\textbf{76.81}}$&$(24.60)_{65.15}$	&$(10.50)_{66.593}$	&$(40.12)_{66.82}$&$(35.45)_{ 66.89}$&$(37.60)_{72.87}$\\
					&GDS3929 &$(41.10)_{73.57}$&$(41.10)_{\textbf{83.78}}$&--			 &$(40.20)_{83.11} $	 	&$(21.60)_{76.97}$	&$(39.90)_{75.82}$&$(41.20)_{ 72.12}$&$(37.90)_{78.76}$\\
					&GDS2546 &$(33.10)_{74.13}$&$(32.70)_{\textbf{83.51}}$&$(27.00)_{77.69}$ &	$(26.40)_{81.25}$&$(17.70)_{80.88}$	&$(35.30)_{73.82}$&$(32.50)_{ 72.98}$&$(32.70)_{81.48}$\\
					&GDS2547 &$(39.40)_{67.31}$&$(32.50)_{73.88}$& $(12.3)_{71.16}$ & $(23.60)_{73.13}$	& $(24.30)_{76.85}$	&$(28.40)_{66.82}$& $(26.60)_{67.35}$&$(33.70)_{\textbf{80.53}}$\\
					&NX100 & $(2.00)_{\textbf{100.00}}$&--				&		--	   		&$(2.00)_{\textbf{100.00}}$	&$(11.00)_{\textbf{100.00}}$	&--&--&$(21.00)_{\textbf{100.00}}$\\
					&NX200 &$(2.40)_{\textbf{100.00}}$& 	--				&		--			&$(2.00)_{\textbf{100.00}}$	 	&$(2.00)_{\textbf{100.00}}$	&--&--&$(12.00)_{\textbf{100.00}}$\\
					\hline

					\multirow{12}{*}{\bf{\rotatebox{45}{RF}}}
					&GDS1615 &$(32.70)_{81.96}$&$(20.20)_{88.24}$&$(22.70)_{\textbf{92.88}}$& $(15.20)_{82.34}$	&$(13.00)_{82.26}$	&$(31.20)_{79.55}$&$(30.80)_{81.08}$&$(31.20)_{85.46}$\\
					&GDS3268 &$(26.50)_{\textbf{87.26}}$&$(41.70)_{86.52}$& -- & $(30.20)_{86.40}$	&$(13.60)_{81.19}$	&$(34.55)_{82.82}$&$(33.50)_{78.87}$& $(32.60)_{86.15}$\\
					&GDS968 &$(44.20)_{79.44}$&$(42.70)_{79.77}$& --			 &$(19.50)_{72.84}$ &$(18.20)_{71.28}$	&$(41.30)_{77.53}$&$(40.73)_{76.42}$&$(38.50)_{\textbf{81.55}}$\\
					&GDS531 & $(23.90)_{63.69}$&$(20.70)_{71.44}$&$(4.70)_{67.82}$&	$(14.80)_{75.48}$	& $(16.40)_{74.67}$	&$(23.60)_{\textbf{77.36}}$&$(21.50)_{73.92}$&$(14.30)_{75.88}$\\
					&GDS2545 & $(31.40)_{79.31}$&$(33.10)_{75.81}$&$(33.10)_{80.64}$&  $(14.80)_{74.16}$		& $(12.00)_{76.05}$	&$(34.20)_{74.82}$&$(33.57)_{75.63}$&$(32.60)_{\textbf{86.78}}$\\
					&GDS1962 & $(29.40)_{72.37}$&$(30.80)_{72.41}$&$(42.1)_{\textbf{78.45}}$&$(21.90)_{69.88}$		&$(13.30)_{63.28}$	&$(32.20)_{69.17}$&$(30.62)_{67.95}$&$(29.30)_{74.32}$\\ 
					&GDS3929 &$(29.10)_{71.94}$&$(28.60)_{73.44}$&--&	$(28.10)_{70.49}$	&$(15.90)_{71.56}$	&$(28.50)_{67.50}$&$(24.10)_{65.13}$&$(18.90)_{66.60}$\\

					&GDS2546 &$(36.30)_{70.53}$&$(34.30)_{75.86}$& $(45.90)_{\textbf{83.09}}$ &	$(25.80)_{77.04}$	&$(18.20)_{78.46}$	&$(36.30)_{72.90}$&$(31.20)_{75.28}$&$(33.30)_{80.31}$\\
					&GDS2547 &$(22.40)_{68.44}$&$(24.80)_{71.68}$& $(32.6)_{\textbf{81.67}}$  &	$(30.00)_{75.85}$	&$(20.50)_{77.10}$	&$(25.40)_{69.70}$&$(24.20)_{71.28}$&$(23.20)_{78.95}$\\
					&NX100 & $(2.00)_{\textbf{100.00}}$&--				&--		  			&$(2.00)_{\textbf{100.00}}$			&$(2.00)_{\textbf{100.00}}$		&--&--&$(22.00)_{\textbf{100.00}}$\\
					&NX200 &$(2.40)_{\textbf{100.00}}$& 	--				&				--&	$(2.00)_{\textbf{100.00}}$			&$(2.00)_{\textbf{100.00}}$	&--&--&$(11.00)_{\textbf{100.00}}$\\
					\hline
				\end{tabular}
			}
			
		\end{adjustbox}
		\label{results_features_accuracy}
	\end{table}
	
	The empty spaces in   Table \ref{results_features_accuracy} under  LARS, HSIC-Lasso, CCM and BCOA's columns simply mean that these methods do not run on those datasets; this is a major shortfall of these methods and it would be interesting to find out why and to what extent LARS, HSIC-Lasso and BCOA fail to run on a dataset. Since the NeuroX datasets have 267,601 features, CCM method requires 1.5 TB of RAM to execute. 
	
	In terms of  accuracy using either of SVM or RF, we can see from Table \ref{results_features_accuracy} that DRPT is at least as good as any of the other seven methods.
	We can further infer that,  in general,  SVM has a better performance than RF on these datasets and 
	SVM requires more features than RF to attain the maximum possible accuracy.

	In Table  \ref{Standard_deviation}, we report  the standard deviation (SD) of the number of selected features and  SD of the classification accuracies over 10 independent runs.   Lower SDs are clearly desirable, which is also an indication of the method's stability with respect to permutation of rows.

	\begin{table}[!ht]
		\centering
		\caption{Superscript is  SD of $\#$ selected features and subscript is the SD of resulting classification accuracies (CA) based on SVM and RF using mRMR, LARS, HSIC-Lasso, Fast-OSFS,  group-SAOLA, CCM, BCOA and DRPT over  10  runs.}
		\centering
		\begin{adjustbox}{width=.92\columnwidth,center}
			\resizebox{\textwidth}{!}{
				\begin{tabular}{c c c c c c c  c c c}
					\hline
			
					\multirow{2}{*}{\bf{\rotatebox{45}{\tiny Classifier}}} &\multirow{2}{*}{\bf{Dataset}} & \multicolumn{4}{c}{$(\text{SD of selected features})_{\text{SD of CA}}$}\\
					&&\bf{mRMR}&\bf{LARS} &\bf{HSIC-Lasso} &\bf{Fast-OSFS} &\bf{group-SAOLA} &\bf{CCM}&\bf{BCOA}&\bf{DRPT} \\
					\hline
					\multirow{12}{*}{\bf{\rotatebox{45}{SVM}}}
					&GDS1615 &$(7.87)_{4.29}$&$(16.54)_{4.95}$&$(8.38)_{4.20}$&	$(6.18)_{4.79}$& $(6.50)_{6.60}$	&$(9.35)_{5.84}$&$(13.60)_{7.43}$&$(7.05)_{\textbf{3.61}}$\\
					&GDS3268 &$(9.35)_{3.58}$&$(5.28)_{3.31}$& -- &$(8.99)_{4.34}$	&${(3.66)}_{3.47}$	&$(6.31)_{4.43}$&$(4.72)_{5.64}$&$(9.49)_{\textbf{1.83}}$\\
					&GDS968 &$(6.20)_{5.12}$&$(8.12)_{4.96}$& -- &$(4.15)_{4.44}$	& $(3.78)_{6.53}$	&$(6.60)_{6.76}$&$(6.50)_{4.12}$&$(7.07)_{\textbf{4.79}}$\\
					&GDS531 & $(17.46)_{4.39}$&$(12.89)_{4.79}$&$(1.45)_{5.74}$&$(9.91)_ {6.30}$	&$(4.77)_{5.71}$	&$(14.05)_{4.79}$&$(14.46)_{3.61}$&$(10.89)_{\textbf{3.02}}$\\
					&GDS2545 & $(13.14)_{3.37}$&$(13.71)_{3.04}$&$(12.13)_{\textbf{2.66}}$&$(9.33)_{6.44}$	&$(6.00)_{6.93}$	&$(12.87)_{5.27}$&$(11.08)_{4.97}$&$(8.97)_{2.79}$\\
					&GDS1962 &$(11.03)_{2.91}$&$(11.55)_{3.68}$&$(15.54)_{4.03}$&$(14.04)_{7.01}$	&$(4.40)_{6.00}$	&$(14.15)_{3.84}$&$(11.04)_{3.57}$&$(10.50)_{\textbf{2.89}}$\\
					&GDS3929 &$(10.51)_{4.64}$&$(11.12)_{3.32}$&-- &$(8.87)_{3.69}$	& $(5.93)_{5.22}$	&$(13.84)_{4.19}$&$(15.53)_{\textbf{3.13}}$&$(9.44)_{3.65}$\\
					&GDS2546 &$(7.96)_{\textbf{2.12}}$&$(10.89)_{4.65}$&$(13.41)_{3.10}$ &$(2.72)_{4.57}$	& $(5.81)_{4.02}$	&$(7.76)_{4.87}$&$(5.25)_{6.82}$&$(10.67)_{4.24}$\\
					&GDS2547 &$(9.36)_{4.42}$&$(9.98)_{3.77}$& $(7.86)_{4.61}$ &$(13.57)_ {4.95}$	& $(8.60)_{4.64}$	&$(8.38)_{5.29}$&$(10.23)_{7.53}$&$(9.48)_{\textbf{3.18}}$\\
					&NX100 & $(00.00)_{\textbf{00.00}}$&--					&--		  			&$(00.00)_{\textbf{00.00}}$	&$(00.00)_{\textbf{00.00}}$	&--&--&$(3.10)_{\textbf{00.00}}$\\
					&NX200 &$(00.71)_{\textbf{00.00}}$& 	--				&				--&$(00.00)_{\textbf{00.00}}$	&$(00.00)_{\textbf{00.00}}$	&--&--&$(2.00)_{\textbf{00.00}}$\\
					\hline
					\multirow{12}{*}{\bf{\rotatebox{45}{RF}}}
					&GDS1615 &$(9.64)_{4.63}$&$(9.27)_{3.44}$&$(7.23)_{\textbf{2.25}}$&$(5.33)_{3.95}$	&$(6.00)_{4.08}$	&$(10.60)_{5.19}$&$(9.60)_{5.29}$&$(9.26)_{3.12}$\\
					&GDS3268 &$(10.24)_{\textbf{2.54}}$&$(6.33)_{2.46}$& -- &$(8.27)_{5.40}$	&$(5.54)_{2.86}$	&$(7.95)_{5.81}$&$(9.04)_{4.13}$&$(8.37)_{2.95}$\\
					&GDS968 &$(4.69)_{4.01}$&$(6.85)_{4.10}$& --			 &$(5.23)_{5.13}$	&$(5.34)_{4.89}$	&$(5.64)_{4.43}$&$(4.22)_{4.03}$&$(5.26)_{\textbf{3.90}}$\\
					&GDS531 & $(10.34)_{3.98}$&$(17.76)_{5.36}$&$(2.11)_{6.57}$&$(7.32)_{9.07}$	&$(6.98)_{6.90}$	&$(9.60)_{6.72}$&$(11.37)_{3.63}$&$(7.38)_{\textbf{2.86}}$\\
					&GDS2545 & $(14.60)_{2.87}$&$(12.08)_{3.28}$&$(15.30)_{\textbf{2.21}}$&$(6.86)_{6.63}$	&$(5.27)_{5.61}$	&$(10.60)_{3.94}$&$(9.98)_{3.71}$&$(9.37)_{2.74}$\\
					&GDS1962 & $(12.00)_{3.20}$&$(15.46)_{\textbf{1.82}}$&$(7.11)_{2.88}$&$(7.29)_{3.54}$	&$(4.87)_{5.48}$	&$(13.52)_{4.82}$&$(10.06)_{5.11}$&$(11.33)_{4.11}$\\
					&GDS3929 &$(11.86)_{2.94}$&$(16.34)_{3.37}$&-- &$(10.84)_{5.31}$	&$(8.94)_{4.14}$	&$(12.60)_{3.96}$&$(9.60)_{4.49}$&$(8.83)_{\textbf{2.48}}$\\
					&GDS2546 &$(11.14)_{3.59}$&$(12.64)_{3.80}$&$(2.73)_{3.33}$ &$(9.39)_{2.36}$	&$(4.02)_{5.25}$	&$(8.73)_{4.17}$&$(10.23)_{4.62}$&$(8.11)_{\textbf{3.15}}$\\
					&GDS2547 &$(15.31)_{4.42}$&$(10.06)_{3.83}$& $(11.19)_{4.16}$ &$(7.63)_{4.58}$	&$(6.15)_{6.03}$	&$(10.28)_{5.19}$&$(11.93)_{4.86}$&$(9.33)_{\textbf{3.25}}$\\
					&NX100 & $(00.00)_{\textbf{00.00}}$&--					&--		  		&$(00.00)_{\textbf{00.00}}$	&$(00.00)_{\textbf{00.00}}$	&--&--&$(2.20)_{\textbf{00.00}}$\\
					&NX200 &$(00.00)_{\textbf{00.00}}$& 	--				&			--&$(00.00)_{\textbf{00.00}}$	&$(00.00)_{\textbf{00.00}}$	&--&--&$(2.30)_{\textbf{00.00}}$\\
					\hline
					
				\end{tabular}
			}
			
		\end{adjustbox}
		\label{Standard_deviation}
	\end{table}
	
	Figure \ref{genomicresults} shows the average classification accuracy results of our DRPT compared to other FS methods using $k$ features and the SVM classifier, where  $k$ is between 10 and 90.  When a FS method returns a subset of $k$ features,  we use SVM to find an optimal $t\leq k$ so that the first $t$ features yield the  best accuracy. Note that  we do not look for the best subset and rather add the features sequentially to find the optimal $t$. Then we calculate the accuracy using the first $t$ features and take the average of these accuracies over 10 runs. Our evaluation metric is consistent across all FS methods.
		
	Figure \ref{genomicresults}  shows the general superiority of the classification accuracy of our proposed model over the other models for the 9 genomic datasets used in our study. We can see a steady increase in classification accuracies of different FS methods as we increase  $k$ from 10 to 50, however the curves usually flat out when $k$ is between 50 and 90. Note that  HSIC-Lasso,  Fast-OSFS, and group-SAOLA models output a  subset of fewer than 60 features.

	We note that the default number of selected features by LARS  is almost the number of samples in a dataset.
	In Table \ref{LS}, we perform a further comparison between LARS and DRPT where we set $k$ to be the default number of features suggested by LARS and we use the classifier to find an optimal subset of size at most $k$. For example, the dataset GDS1615, has 127 samples in total. Since we take approximately 70\% of samples for FS, the suggested number of features by LARS is  $k=87$.

	\begin{table}[!ht]
		\centering
		\caption{Superscript is average number of selected features and subscript is resulting classification accuracies (CA) based on SVM and RF using LARS Suggestion (LS) for 10 independent runs of DRPT and LARS.}
		\centering
		\begin{adjustbox}{width=.6\columnwidth,center}
			\resizebox{\textwidth}{!}{
				\begin{tabular}{c c c c c  }
					\hline
					\multirow{2}{*}{\bf{\rotatebox{45}{\tiny Classifier}}} &\multirow{2}{*}{\bf{Dataset}} & \multicolumn{3}{c}{$(\text{\# of selected features})_{\text{classification accuracy}}$}\\
					&& \textbf{\# of LS}&\bf{DRPT}&\bf{LARS}  \\\hline
					\multirow{12}{*}{\bf{\rotatebox{45}{SVM}}}
					&GDS1615 &$87$&$(69.90)_{95.89}$&$(62.2)_{93.99}$\\
					&GDS3268 & $140$&$(94.50)_{95.13}$ &$(103.30)_{95.15}$\\
					&GDS968 &  $118$& $(90.50)_{86.93}$ &$(68.30)_{87.20}$\\
					&GDS531 &$120$&$(92.80)_{83.79}$&$(51.15)_{81.07}$\\
					&GDS2545 &$118$&$(63.00)_{84.33}$&$(58.70)_{80.28}$\\
					&GDS1962 &$124$&$(75.47)_{75.19}$&$(44.3)_{77.43}$\\
					&GDS3929 &$127$&$(93.57)_{89.75}$ &$(78.70)_{86.77}$\\
					
					&GDS2546 &$115$&$(89.12)_{85.92}$ &$(61.90)_{84.31}$\\
					&GDS2547 & $113$&$(83.94)_{85.14}$ &$(67.30)_{77.67}$\\
					\hline
					\multirow{12}{*}{\bf{\rotatebox{45}{RF}}}
					&GDS1615 &$87$&$(57.76)_{89.83}$&$(54.00)_{90.23}$\\
					&GDS3268 & $140$&$(84.4)_{89.35}$ &$(87.20)_{90.67}$\\
					&GDS968 & $118$&$(91.30)_{87.20}$	&$(85.30)_{85.14}$\\
					&GDS531 &$120$&$(26.50)_{76.23}$&$(25.60)_{75.54}$\\
					&GDS2545 &$118$&$(73.94)_{89.72}$&$(75.50)_{78.65}$\\
					&GDS1962 &$124$&$(89.75)_{75.00}$&$(70.90)_{74.68}$\\ 
					&GDS3929 &$127$&$(52.10)_{70.51}$&$(22.90)_{74.32}$\\
					
					&GDS2546 &$115$& $(47.80)_{82.49}$ & $(53.10)_{77.93}$\\
					&GDS2547 &$113$& $(53.80)_{78.86}$  & $(28.00)_{73.87}$\\
					\hline
					
				\end{tabular}
			}
		\end{adjustbox}
		\label{LS}
	\end{table}

	If we look at the performance of LARS just based on its default number of features, we note that  
	CA of LARS significantly drops.  This, in particular, suggests that LARS does not select an optimal subset of features.

	\pgfplotstableread[row sep=\\,col sep=&]{
	datasets & mrmr & lars   &HSIC-Lasso&Fast   &Saola & CCM  & BCOA  & BCOA_new&DRPT  \\  
	10          & 65.14 & 81.34&    82.99    & 81.23&80.72&63.62  & 63.14  & 64.10       &83.52\\
	20          &  72.10& 82.54&    84.90      &84.31 &83.13&66.65&71.72   & 73.90       &84.96\\
	30          & 83.25 & 83.86&    87.25    &84.31 &83.13 &67.19 & 77.18  & 79.19       &85.48\\
	40          &  86.28& 87.58&    91.35    &           &         &72.06 & 79.95  & 82.55       &88.95\\
	50          & 87.37 &91.67 &                  &          &         &80.82& 84.31   & 84.90       &91.89\\
	60          &  88.02& 92.62&                  &          &         &81.96&84.52    & 85.08       &92.96\\
	70          & 90.25 & 93.16&                  &          &         &81.50& 85.64   & 85.97       &95.48\\
	80          &  92.10& 93.58&                  &          &         &82.49& 87.75   & 87.12       &95.87\\
	90          & 94.10 &93.99 &                  &          &         &82.92& 88.98   & 88.50       &95.87\\
}\mydata

\definecolor{eggplant}{rgb}{0.38, 0.25, 0.32}
\definecolor{byzantium}{rgb}{0.44, 0.16, 0.39}
\definecolor{tyrianpurple}{rgb}{0.4, 0.01, 0.24}
\definecolor{internationalorange}{rgb}{1.0, 0.31, 0.0}
\definecolor{laserlemon}{rgb}{1.0, 1.0, 0.13}
\begin{figure}
	\centering 
	
	\subfigure{
		\begin{tikzpicture} 	
		\begin{axis}[
		title style={font=\tiny,yshift=-1.8ex,},
		title=GDS1615,
		width  = 0.37*\textwidth,
		height =4 cm,
		major x tick style = transparent,
		ymajorgrids = true,
		label style={font=\tiny},
		ylabel = {\tiny CA},
		ytick={65,70,75,80,85,90,95},
		symbolic x coords={10,20,30,40,50,60,70,80,90},
		xticklabels={10,20,30,40,50,60,70,80,90},
		tick label style={font=\tiny},
		xtick=data,
		scaled y ticks = false,
		enlarge x limits=0.06,
		ymin=60,
		legend columns=2,
		legend cell align=left,
		legend style={nodes={scale=0.6, transform shape},at={(axis cs:90,111)},anchor=north east,font=\tiny}
		]
		\pgfmathtruncatemacro{\PlotNum}{10}
		\addplot+[yshift=\PlotNum*\pgflinewidth,mark options={fill=blue!50!white}] table[x=datasets,y=mrmr]{\mydata};
		\pgfmathtruncatemacro{\PlotNum}{\PlotNum-1}
		\addplot+[yshift=\pgflinewidth,mark options={fill=red!50!white}] table[x=datasets,y=lars]{\mydata};
		\pgfmathtruncatemacro{\PlotNum}{\PlotNum-1}
		\addplot +[yshift=\pgflinewidth,mark options={fill=brown!50!white}] table[x=datasets,y=HSIC-Lasso]{\mydata};
		\pgfmathtruncatemacro{\PlotNum}{\PlotNum-1}
		\addplot+[yshift=\pgflinewidth,mark options={fill=black!50!white}] table[x=datasets,y=Fast]{\mydata};
		\pgfmathtruncatemacro{\PlotNum}{\PlotNum-1}
		\addplot +[yshift=\pgflinewidth,color=tyrianpurple!50!black ,mark options={fill=tyrianpurple!50!white}] table[x=datasets,y=Saola]{\mydata};
		\pgfmathtruncatemacro{\PlotNum}{\PlotNum-1}
		\addplot+[yshift=\pgflinewidth,color=laserlemon!50!black,solid ,mark options={fill=laserlemon!50!white}] table[x=datasets,y=CCM]{\mydata};
		\pgfmathtruncatemacro{\PlotNum}{\PlotNum-1}
		\addplot +[yshift=\pgflinewidth,pink!50!black,solid,mark options={fill=pink!50!white}] table[x=datasets,y=BCOA_new]{\mydata};
		\pgfmathtruncatemacro{\PlotNum}{\PlotNum-1}
		\addplot +[yshift=\pgflinewidth,green!50!black,solid,mark options={fill=green!50!white}] table[x=datasets,y=DRPT]{\mydata};
		\end{axis}
		\end{tikzpicture}

	}
	\pgfplotstableread[row sep=\\,col sep=&]{
	datasets & mrmr & lars   &HSIC-Lasso&Fast   &Saola & CCM  & BCOA  &BCOA_new &DRPT  \\  
	10          & 77.14 & 77.34&    0            & 69.23&68.72&72.62 & 58.26  & 62.55		&77.52\\
	20          &  81.10& 81.54&      1           &79.36 &78.69&77.65& 61.45  & 63.80		&81.96\\
	30          & 83.25 & 83.86&                  &83.06 &79.12&82.19& 63.29  & 66.91 		 &83.48\\
	40          &  84.28& 86.58&     	          & 86.67& 84.13&83.06& 68.43 & 72.25        &86.95\\
	50          & 85.69 &89.62 &                  & 87.89& 84.13&85.82& 70.12 & 73.55        &90.45\\
	60          &  87.02& 89.62&                  &88.12 &         &86.96& 71.14  & 75.67        &89.96\\
	70          & 88.25 & 90.16&                  &          &         &87.50& 73.18 & 76.90         &89.48\\
	80          &  89.10& 89.58&                  &          &         &86.49& 75.19 & 78.21         &90.87\\
	90          & 89.10 &89.34 &                  &          &         &87.92& 77.58& 79.98          &90.87\\
}\mydatab
	\subfigure{
		\begin{tikzpicture} 	
		\begin{axis}[
		title style={font=\tiny,yshift=-1.8ex,},
		title=GDS3268,
		width  = 0.37*\textwidth,
		height = 4 cm,
		major x tick style = transparent,
		ymajorgrids = true,
		label style={font=\tiny},
		ytick={60,65,70,75,80,85,90},
		symbolic x coords={10,20,30,40,50,60,70,80,90},
		xticklabels={10,20,30,40,50,60,70,80,90},
		tick label style={font=\tiny},
		xtick=data,
		scaled y ticks = false,
		enlarge x limits=0.06,
		ymin=55,
		legend columns=2,
		legend cell align=left,
		legend style={nodes={scale=0.6, transform shape},at={(axis cs:90,118)},anchor=north east,font=\tiny}
		]
		\pgfmathtruncatemacro{\PlotNum}{10}
		\addplot+[yshift=\PlotNum*\pgflinewidth,mark options={fill=blue!50!white}] table[x=datasets,y=mrmr]{\mydatab};
		\pgfmathtruncatemacro{\PlotNum}{\PlotNum-1}
		\addplot+[yshift=\pgflinewidth,mark options={fill=red!50!white}] table[x=datasets,y=lars]{\mydatab};
		\pgfmathtruncatemacro{\PlotNum}{\PlotNum-1}
		\addplot +[yshift=\pgflinewidth,mark options={fill=brown!50!white}] table[x=datasets,y=HSIC-Lasso]{\mydatab};
		\pgfmathtruncatemacro{\PlotNum}{\PlotNum-1}
		\addplot+[yshift=\pgflinewidth,mark options={fill=black!50!white}] table[x=datasets,y=Fast]{\mydatab};
		\pgfmathtruncatemacro{\PlotNum}{\PlotNum-1}
		\addplot +[yshift=\pgflinewidth,color=tyrianpurple!50!black ,mark options={fill=tyrianpurple!50!white}] table[x=datasets,y=Saola]{\mydatab};
		\pgfmathtruncatemacro{\PlotNum}{\PlotNum-1}
		\addplot+[yshift=\pgflinewidth,color=laserlemon!50!black,solid ,mark options={fill=laserlemon!50!white}] table[x=datasets,y=CCM]{\mydatab};
		\pgfmathtruncatemacro{\PlotNum}{\PlotNum-1}
		\addplot +[yshift=\pgflinewidth,pink!50!black,solid,mark options={fill=pink!50!white}] table[x=datasets,y=BCOA_new]{\mydatab};
		\pgfmathtruncatemacro{\PlotNum}{\PlotNum-1}
		\addplot +[yshift=\pgflinewidth,green!50!black,solid,mark options={fill=green!50!white}] table[x=datasets,y=DRPT]{\mydatab};
		\legend{mRMR,Lars,HSIC-Lasso,OSFS,SAOLA,CCM,BCOA,DRPT}
		\end{axis}
		\end{tikzpicture}

	}
	\pgfplotstableread[row sep=\\,col sep=&]{
	datasets & mrmr & lars   &HSIC-Lasso&Fast   &Saola & new  & BCOA & BCOA_new& DRPT  \\  
	10          & 52.47 & 63.65&                  & 60.23&61.72&57.62& 50.12 & 52.90        &65.52\\
	20          &  65.77& 71.54&                  &69.36 &66.69&65.65& 54.18 & 58.35        &74.96\\
	30          & 74.25 & 79.86&                  & 72.41&70.53&72.19& 65.18 & 66.19        &76.48\\
	40          &  76.88& 82.58&     	          &          &         &78.06& 72.18 & 73.12        &80.35\\
	50          & 80.37 &83.34 &                  &          &         &78.82& 76.19 & 76.19        &81.09\\
	60          &  87.77& 85.54&                  &          &         &80.96& 78.57 & 78.20        &83.96\\
	70          & 87.25 & 86.86&                  &          &         &83.50& 78.99 & 78.50        & 85.48\\
	80          &  88.88& 87.58&                  &          &         &85.49& 80.18 & 80.55        &87.87\\
	90          & 90.37 &88.34 &                  &          &         &85.92& 80.18 & 80.22        &88.09\\
}\mydatac
	\subfigure{
		\begin{tikzpicture} 	
		\begin{axis}[
		title style={font=\tiny,yshift=-1.8ex,},
		title=GDS968,
		width  = 0.37*\textwidth,
		height = 4 cm,
		major x tick style = transparent,
		ymajorgrids = true,
		label style={font=\tiny},
		ytick={55,50,60,65,70,75,80,85,90},
		symbolic x coords={10,20,30,40,50,60,70,80,90},
		xticklabels={10,20,30,40,50,60,70,80,90},
		tick label style={font=\tiny},
		xtick=data,
		scaled y ticks = false,
		enlarge x limits=0.06,
		ymin=50,
		legend columns=1,
		legend cell align=left,
		legend style={nodes={scale=.8, transform shape},at={(axis cs:0,83)},anchor=south west,font=\tiny}
		]
	\pgfmathtruncatemacro{\PlotNum}{0}
	\addplot+[yshift=\PlotNum*\pgflinewidth,mark options={fill=blue!50!white}] table[x=datasets,y=mrmr]{\mydatac};
	\pgfmathtruncatemacro{\PlotNum}{\PlotNum+1}
	\addplot+[yshift=\pgflinewidth,mark options={fill=red!50!white}] table[x=datasets,y=lars]{\mydatac};
	\pgfmathtruncatemacro{\PlotNum}{\PlotNum+1}
	\addplot +[yshift=\pgflinewidth,mark options={fill=brown!50!white}] table[x=datasets,y=HSIC-Lasso]{\mydatac};
	\pgfmathtruncatemacro{\PlotNum}{\PlotNum+1}
	\addplot+[yshift=\pgflinewidth,mark options={fill=black!50!white}] table[x=datasets,y=Fast]{\mydatac};
	\pgfmathtruncatemacro{\PlotNum}{\PlotNum+1}
	\addplot +[yshift=\pgflinewidth,color=tyrianpurple!50!black ,mark options={fill=tyrianpurple!50!white}] table[x=datasets,y=Saola]{\mydatac};
	\pgfmathtruncatemacro{\PlotNum}{\PlotNum+1}
	\addplot+[yshift=\pgflinewidth,color=laserlemon!50!black,solid ,mark options={fill=laserlemon!50!white}] table[x=datasets,y=new]{\mydatac};
	\pgfmathtruncatemacro{\PlotNum}{\PlotNum-1}
	\addplot +[yshift=\pgflinewidth,pink!50!black,solid,mark options={fill=pink!50!white}] table[x=datasets,y=BCOA_new]{\mydatac};
	\pgfmathtruncatemacro{\PlotNum}{\PlotNum+1}
	\addplot +[yshift=\pgflinewidth,green!50!black,solid,mark options={fill=green!50!white}] table[x=datasets,y=DRPT]{\mydatac};
		\end{axis}
		\end{tikzpicture}
		
	}

	\pgfplotstableread[row sep=\\,col sep=&]{
	datasets & mrmr & lars   &HSIC-Lasso&Fast   &Saola & CCM  & BCOA& BCOA_new & DRPT  \\  
	10          & 60.14 & 70.34&    67.93     & 68.23&69.72&73.62& 61.47& 62.50          &71.52\\
	20          &  61.10& 72.54&                  &73.36 &74.69&75.65& 68.78& 69.84          &73.96\\
	30          & 64.25 & 76.86&                  &75.06 &77.70&78.19& 71.34& 72.22          &76.48\\
	40          &  67.28& 77.58&     	          & 77.43&         &80.06& 73.48&74.01           &76.95\\
	50          & 69.69 &79.62 &                  &          &         &80.82& 74.17& 74.08          &77.16\\
	60          &  70.02& 79.62&                  &          &         &81.96&74.46 & 74.25          &78.96\\
	70          & 70.25 & 80.31&                  &          &         &82.50& 75.15& 74.87          &78.48\\
	80          &  72.10& 80.31&                  &          &         &80.49& 76.48& 75.90           &80.87\\
	90          & 74.10 &80.31 &                  &          &         &82.92&76.48 & 75.68            &83.87\\
}\mydatab
\subfigure{
	\begin{tikzpicture} 	
	\begin{axis}[
	title style={font=\tiny,yshift=-1.8ex,},
	title=GDS531,
	width  = 0.37*\textwidth,
	height = 4 cm,
	major x tick style = transparent,
	ymajorgrids = true,
	label style={font=\tiny},
	ylabel = {\tiny CA},
	ytick={65,70,75,80,85,90},
	symbolic x coords={10,20,30,40,50,60,70,80,90},
	xticklabels={10,20,30,40,50,60,70,80,90},
	tick label style={font=\tiny},
	xtick=data,
	scaled y ticks = false,
	enlarge x limits=0.06,
	ymin=60,
	legend columns=2,
	legend cell align=left,
	legend style={nodes={scale=0.6, transform shape},at={(axis cs:90,111)},anchor=north east,font=\tiny}
	]
	\pgfmathtruncatemacro{\PlotNum}{10}
	\addplot+[yshift=\PlotNum*\pgflinewidth,mark options={fill=blue!50!white}] table[x=datasets,y=mrmr]{\mydatab};
	\pgfmathtruncatemacro{\PlotNum}{\PlotNum-1}
	\addplot+[yshift=\pgflinewidth,mark options={fill=red!50!white}] table[x=datasets,y=lars]{\mydatab};
	\pgfmathtruncatemacro{\PlotNum}{\PlotNum-1}
	\addplot +[yshift=\pgflinewidth,mark options={fill=brown!50!white}] table[x=datasets,y=HSIC-Lasso]{\mydatab};
	\pgfmathtruncatemacro{\PlotNum}{\PlotNum-1}
	\addplot+[yshift=\pgflinewidth,mark options={fill=black!50!white}] table[x=datasets,y=Fast]{\mydatab};
	\pgfmathtruncatemacro{\PlotNum}{\PlotNum-1}
	\addplot +[yshift=\pgflinewidth,color=tyrianpurple!50!black ,mark options={fill=tyrianpurple!50!white}] table[x=datasets,y=Saola]{\mydatab};
	\pgfmathtruncatemacro{\PlotNum}{\PlotNum-1}
	\addplot+[yshift=\pgflinewidth,color=laserlemon!50!black,solid ,mark options={fill=laserlemon!50!white}] table[x=datasets,y=CCM]{\mydatab};
	\pgfmathtruncatemacro{\PlotNum}{\PlotNum-1}
	\addplot +[yshift=\pgflinewidth,pink!50!black,solid,mark options={fill=pink!50!white}] table[x=datasets,y=BCOA_new]{\mydatab};
	\pgfmathtruncatemacro{\PlotNum}{\PlotNum-1}
	\addplot +[yshift=\pgflinewidth,green!50!black,solid,mark options={fill=green!50!white}] table[x=datasets,y=DRPT]{\mydatab};
	\end{axis}
	\end{tikzpicture}
	}
	\pgfplotstableread[row sep=\\,col sep=&]{
datasets & mrmr & lars   &HSIC-Lasso&Fast   &Saola & CCM  & BCOA& BCOA_new&DRPT  \\  
10          & 59.14 & 67.34&    69.34      & 69.23&65.72&62.62& 61.34& 63.15        &74.52\\
20          &  65.10& 74.54&     73.24    &73.36 &70.69&65.65& 66.54 & 68.19        &81.96\\
30          & 70.25 & 77.86&     74.39    &74.22 &75.12&68.19& 67.85 & 69.42        &82.48\\
40          &  73.28& 78.58&    75.54     & 74.95&         &68.06& 71.78 & 72.90        &83.23\\
50          & 75.90 &79.62 &    76.40     &          &         &70.82& 74.95 & 75.40        &83.23\\
60          &  76.02& 79.62&    77.56     &          &         &71.96&74.72 & 75.40         &84.96\\
70          & 78.25 & 79.62&                  &          &         &72.50&74.72 & 76              &85.48\\
80          &  78.25& 79.62&                  &          &         &73.49&75.21 &    76.52      &84.87\\
90          & 77.10 &80.34 &                  &          &         &73.92&73.17 & 77.01         &84.87\\
}\mydatab
\subfigure{
\begin{tikzpicture} 	
\begin{axis}[
title style={font=\tiny,yshift=-1.8ex,},
title=GDS2545,
width  = 0.37*\textwidth,
height = 4 cm,
major x tick style = transparent,
ymajorgrids = true,
label style={font=\tiny},
ytick={60,65,70,75,80,85,90},
symbolic x coords={10,20,30,40,50,60,70,80,90},
xticklabels={10,20,30,40,50,60,70,80,90},
tick label style={font=\tiny},
xtick=data,
scaled y ticks = false,
enlarge x limits=0.06,
ymin=55,
legend columns=2,
legend cell align=left,
legend style={nodes={scale=0.6, transform shape},at={(axis cs:90,111)},anchor=north east,font=\tiny}
]
\pgfmathtruncatemacro{\PlotNum}{10}
\addplot+[yshift=\PlotNum*\pgflinewidth,mark options={fill=blue!50!white}] table[x=datasets,y=mrmr]{\mydatab};
\pgfmathtruncatemacro{\PlotNum}{\PlotNum-1}
\addplot+[yshift=\pgflinewidth,mark options={fill=red!50!white}] table[x=datasets,y=lars]{\mydatab};
\pgfmathtruncatemacro{\PlotNum}{\PlotNum-1}
\addplot +[yshift=\pgflinewidth,mark options={fill=brown!50!white}] table[x=datasets,y=HSIC-Lasso]{\mydatab};
\pgfmathtruncatemacro{\PlotNum}{\PlotNum-1}
\addplot+[yshift=\pgflinewidth,mark options={fill=black!50!white}] table[x=datasets,y=Fast]{\mydatab};
\pgfmathtruncatemacro{\PlotNum}{\PlotNum-1}
\addplot +[yshift=\pgflinewidth,color=tyrianpurple!50!black ,mark options={fill=tyrianpurple!50!white}] table[x=datasets,y=Saola]{\mydatab};
\pgfmathtruncatemacro{\PlotNum}{\PlotNum-1}
\addplot+[yshift=\pgflinewidth,color=laserlemon!50!black,solid ,mark options={fill=laserlemon!50!white}] table[x=datasets,y=CCM]{\mydatab};
\pgfmathtruncatemacro{\PlotNum}{\PlotNum-1}
\addplot +[yshift=\pgflinewidth,pink!50!black,solid,mark options={fill=pink!50!white}] table[x=datasets,y=BCOA_new]{\mydatab};
\pgfmathtruncatemacro{\PlotNum}{\PlotNum-1}
\addplot +[yshift=\pgflinewidth,green!50!black,solid,mark options={fill=green!50!white}] table[x=datasets,y=DRPT]{\mydatab};
\end{axis}
\end{tikzpicture}
	}
	\pgfplotstableread[row sep=\\,col sep=&]{
datasets & mrmr & lars   &HSIC-Lasso&Fast   &Saola & new  & BCOA& BCOA_new&DRPT  \\  
10          & 55.47 & 62.65&   66.12      & 59.23&56.72&56.62& 56.68& 58.88        &62.52\\
20          &  59.77& 69.54&   72.34      &61.36 &61.69&60.65& 60.56& 63.03        &67.96\\
30          & 62.25 & 73.86&   74.52      & 63.41&66.59&64.19& 63.74& 64.92        &69.48\\
40          &  63.88& 74.58&   76.16      & 64.74&         &65.06& 64.68& 65.72        &70.35\\
50          & 65.37 &76.34 &   76.40      & 65.15&         &66.82& 65.45&  66.89        &72.78\\
60          &  66.77& 76.34&   77.66     &          &         &66.96& 68.13 &  68.01        &74.99\\
70          & 67.25 & 77.58&                 &          &         &68.50& 70.18 & 71.62         &75.48\\
80          &  68.88& 77.58&                  &          &         &67.49& 72.34& 71.99         &76.87\\
90          & 69.37 &77.58 &                  &          &         &68.92& 69.89& 73.18         &77.09\\
}\mydatac
\subfigure{
\begin{tikzpicture} 	
\begin{axis}[
title style={font=\tiny,yshift=-1.8ex,},
title=GDS1962,
width  = 0.37*\textwidth,
height = 4 cm,
major x tick style = transparent,
ymajorgrids = true,
label style={font=\tiny},
ytick={60,65,70,75,80,85,90},
symbolic x coords={10,20,30,40,50,60,70,80,90},
xticklabels={10,20,30,40,50,60,70,80,90},
tick label style={font=\tiny},
xtick=data,
scaled y ticks = false,
enlarge x limits=0.06,
ymin=55,
legend columns=1,
legend cell align=left,
legend style={nodes={scale=.8, transform shape},at={(axis cs:0,83)},anchor=south west,font=\tiny}
]
\pgfmathtruncatemacro{\PlotNum}{0}
\addplot+[yshift=\PlotNum*\pgflinewidth,mark options={fill=blue!50!white}] table[x=datasets,y=mrmr]{\mydatac};
\pgfmathtruncatemacro{\PlotNum}{\PlotNum+1}
\addplot+[yshift=\pgflinewidth,mark options={fill=red!50!white}] table[x=datasets,y=lars]{\mydatac};
\pgfmathtruncatemacro{\PlotNum}{\PlotNum+1}
\addplot +[yshift=\pgflinewidth,mark options={fill=brown!50!white}] table[x=datasets,y=HSIC-Lasso]{\mydatac};
\pgfmathtruncatemacro{\PlotNum}{\PlotNum+1}
\addplot+[yshift=\pgflinewidth,mark options={fill=black!50!white}] table[x=datasets,y=Fast]{\mydatac};
\pgfmathtruncatemacro{\PlotNum}{\PlotNum+1}
\addplot +[yshift=\pgflinewidth,color=tyrianpurple!50!black ,mark options={fill=tyrianpurple!50!white}] table[x=datasets,y=Saola]{\mydatac};
\pgfmathtruncatemacro{\PlotNum}{\PlotNum+1}
\addplot+[yshift=\pgflinewidth,color=laserlemon!50!black,solid ,mark options={fill=laserlemon!50!white}] table[x=datasets,y=new]{\mydatac};
\pgfmathtruncatemacro{\PlotNum}{\PlotNum-1}
\addplot +[yshift=\pgflinewidth,pink!50!black,solid,mark options={fill=pink!50!white}] table[x=datasets,y=BCOA_new]{\mydatac};
\pgfmathtruncatemacro{\PlotNum}{\PlotNum+1}
\addplot +[yshift=\pgflinewidth,green!50!black,solid,mark options={fill=green!50!white}] table[x=datasets,y=DRPT]{\mydatac};

\end{axis}
\end{tikzpicture}
		
	}
	\pgfplotstableread[row sep=\\,col sep=&]{
	datasets & mrmr & lars   &HSIC-Lasso&Fast   &Saola & CCM  & BCOA& BCOA_new&DRPT  \\  
	10          & 55.23 & 64.34&    67.93     & 65.23&67.72&60.62& 54.47 & 58.04	    &66.52\\
	20          &  65.10& 74.54&                  &76.36 &74.69&66.65& 58.78 & 62.35        &68.96\\
	30          & 68.25 & 78.86&                  &79.06 &77.70&69.19& 62.34 & 66.08        &70.48\\
	40          &  71.28& 80.58&     	          & 81.43&76.90&73.06& 63.48 & 68.22        &76.95\\
	50          & 73.69 &83.78 &                  & 83.11 &         &75.82& 68.17& 72.12        &78.76\\
	60          &  74.02& 84.62&                  & 81.55&         &75.96 &68.46 & 72.94        &80.96\\
	70          & 72.25 & 84.31&                  &          &         &76.50& 69.15 & 74.05        &81.48\\
	80          &  75.10& 85.31&                  &          &         &76.49& 69.48 & 74.88        &83.87\\
	90          & 76.10 &85.31 &                  &          &         &76.92&70.48  &76.11         &85.87\\
}\mydata
\subfigure{
	\begin{tikzpicture} 	
		\begin{axis}[
			title style={font=\tiny,yshift=-1.8ex,},
			title=GDS3929,
			width  = 0.37*\textwidth,
			height = 4 cm,
			major x tick style = transparent,
			ymajorgrids = true,
			label style={font=\tiny},
			ylabel = {\tiny CA},
			xlabel={\# Features},
			ytick={55,60,65,70,75,80,85,90},
			symbolic x coords={10,20,30,40,50,60,70,80,90},
			xticklabels={10,20,30,40,50,60,70,80,90},
			tick label style={font=\tiny},
			xtick=data,
			scaled y ticks = false,
			enlarge x limits=0.06,
			ymin=50,
			]
			\pgfmathtruncatemacro{\PlotNum}{10}
			\addplot+[yshift=\PlotNum*\pgflinewidth,mark options={fill=blue!50!white}] table[x=datasets,y=mrmr]{\mydata};
			\pgfmathtruncatemacro{\PlotNum}{\PlotNum-1}
			\addplot+[yshift=\pgflinewidth,mark options={fill=red!50!white}] table[x=datasets,y=lars]{\mydata};
			\pgfmathtruncatemacro{\PlotNum}{\PlotNum-1}
			\addplot +[yshift=\pgflinewidth,mark options={fill=brown!50!white}] table[x=datasets,y=HSIC-Lasso]{\mydata};
			\pgfmathtruncatemacro{\PlotNum}{\PlotNum-1}
			\addplot+[yshift=\pgflinewidth,mark options={fill=black!50!white}] table[x=datasets,y=Fast]{\mydata};
			\pgfmathtruncatemacro{\PlotNum}{\PlotNum-1}
			\addplot +[yshift=\pgflinewidth,color=tyrianpurple!50!black ,mark options={fill=tyrianpurple!50!white}] table[x=datasets,y=Saola]{\mydata};
			\pgfmathtruncatemacro{\PlotNum}{\PlotNum-1}
			\addplot+[yshift=\pgflinewidth,color=laserlemon!50!black,solid ,mark options={fill=laserlemon!50!white}] table[x=datasets,y=CCM]{\mydata};
			\pgfmathtruncatemacro{\PlotNum}{\PlotNum-1}
			\addplot +[yshift=\pgflinewidth,pink!50!black,solid,mark options={fill=pink!50!white}] table[x=datasets,y=BCOA_new]{\mydata};
			\pgfmathtruncatemacro{\PlotNum}{\PlotNum-1}
			\addplot +[yshift=\pgflinewidth,green!50!black,solid,mark options={fill=green!50!white}] table[x=datasets,y=DRPT]{\mydata};
		\end{axis}
	\end{tikzpicture}
}
	\pgfplotstableread[row sep=\\,col sep=&]{
	datasets & mrmr & lars   &HSIC-Lasso&Fast   &Saola & CCM  & BCOA& BCOA_new&DRPT  \\  
	10          & 57.23 & 70.34&    67.61     & 69.23&72.72&52.62& 51.47& 53.64        &78.52\\
	20          &  67.10& 76.54&    75.52     &75.36 &76.69&60.65& 58.78& 60.97        &78.96\\
	30          & 71.25 & 79.86&    77.01     &81.06 &77.70&65.19& 62.34& 64.49        &79.48\\
	40          &  71.28& 80.58&   77.50      & 81.25&80.88&70.06& 68.48& 69.31        &81.65\\
	50          &  74.69&83.51 &    77.69     &          &         &73.82& 72.35& 72.98        &81.48\\
	60          &  76.02& 83.66&    78.31     &          &         &74.96&73.46& 73.55         &83.96\\
	70          & 77.25 & 84.11&                  &          &         &75.50& 73.15& 74.72        &83.48\\
	80          &  78.10& 84.11&                  &          &         &75.49& 74.48& 74.55        &84.87\\
	90          & 79.10 &84.25 &                  &          &         &75.92&75.48 & 76             &84.87\\
}\mydata
\subfigure{
	\begin{tikzpicture} 	
	\begin{axis}[
	title style={font=\tiny,yshift=-1.8ex,},
	title=GDS2546,
	width  = 0.37*\textwidth,
	height = 4 cm,
	major x tick style = transparent,
	ymajorgrids = true,
	label style={font=\tiny},
	xlabel={\# Features},
	ytick={55,60,65,70,75,80,85,90},
	symbolic x coords={10,20,30,40,50,60,70,80,90},
	xticklabels={10,20,30,40,50,60,70,80,90},
	tick label style={font=\tiny},
	xtick=data,
	scaled y ticks = false,
	enlarge x limits=0.06,
	ymin=50,
	]
	\pgfmathtruncatemacro{\PlotNum}{10}
	\addplot+[yshift=\PlotNum*\pgflinewidth,mark options={fill=blue!50!white}] table[x=datasets,y=mrmr]{\mydata};
	\pgfmathtruncatemacro{\PlotNum}{\PlotNum-1}
	\addplot+[yshift=\pgflinewidth,mark options={fill=red!50!white}] table[x=datasets,y=lars]{\mydata};
	\pgfmathtruncatemacro{\PlotNum}{\PlotNum-1}
	\addplot +[yshift=\pgflinewidth,mark options={fill=brown!50!white}] table[x=datasets,y=HSIC-Lasso]{\mydata};
	\pgfmathtruncatemacro{\PlotNum}{\PlotNum-1}
	\addplot+[yshift=\pgflinewidth,mark options={fill=black!50!white}] table[x=datasets,y=Fast]{\mydata};
	\pgfmathtruncatemacro{\PlotNum}{\PlotNum-1}
	\addplot +[yshift=\pgflinewidth,color=tyrianpurple!50!black ,mark options={fill=tyrianpurple!50!white}] table[x=datasets,y=Saola]{\mydata};
	\pgfmathtruncatemacro{\PlotNum}{\PlotNum-1}
	\addplot+[yshift=\pgflinewidth,color=laserlemon!50!black,solid ,mark options={fill=laserlemon!50!white}] table[x=datasets,y=CCM]{\mydata};
	\pgfmathtruncatemacro{\PlotNum}{\PlotNum-1}
	\addplot +[yshift=\pgflinewidth,pink!50!black,solid,mark options={fill=pink!50!white}] table[x=datasets,y=BCOA_new]{\mydata};
	\pgfmathtruncatemacro{\PlotNum}{\PlotNum-1}
	\addplot +[yshift=\pgflinewidth,green!50!black,solid,mark options={fill=green!50!white}] table[x=datasets,y=DRPT]{\mydata};
	\end{axis}
	\end{tikzpicture}
}
	\pgfplotstableread[row sep=\\,col sep=&]{
	datasets & mrmr & lars   &HSIC-Lasso&Fast   &Saola & CCM  & BCOA& DRPT  \\  
	10          & 54.23 & 62.34&    67.61     & 66.93&66.72&53.62& 53.47&69.52\\
	20          &  60.10& 69.54&    71.16     &71.36 &72.69&58.65& 59.78&73.96\\
	30          & 62.25 & 71.86&                  &72.06 &74.70&62.19& 63.34&75.48\\
	40          &  64.28& 73.58&                  & 73.13&76.85&62.06& 64.48&80.12\\
	50          &  67.31&73.88 &                  & 73.13&         &66.82& 67.35&80.53\\
	60          &  69.02& 74.66&                  & 75.23&         &67.96&68.46&79.96\\
	70          & 73.25 & 75.11&                  &          &         &69.50& 71.15&80.48\\
	80          &  72.10& 76.11&                  &          &         &71.49& 73.48&80.87\\
	90          & 74.10 &77.25 &                  &          &         &73.92&75.48 & 82.87\\
}\mydata
\subfigure{
	\begin{tikzpicture} 	
	\begin{axis}[
	title style={font=\tiny,yshift=-1.8ex,},
	title=GDS2547,
	width  = 0.37*\textwidth,
	height = 4 cm,
	major x tick style = transparent,
	ymajorgrids = true,
	label style={font=\tiny},
	xlabel={\# Features},
	ytick={55,60,65,70,75,80,85,90},
	symbolic x coords={10,20,30,40,50,60,70,80,90},
	xticklabels={10,20,30,40,50,60,70,80,90},
	tick label style={font=\tiny},
	xtick=data,
	scaled y ticks = false,
	enlarge x limits=0.06,
	ymin=50,
	]
	\pgfmathtruncatemacro{\PlotNum}{10}
	\addplot+[yshift=\PlotNum*\pgflinewidth,mark options={fill=blue!50!white}] table[x=datasets,y=mrmr]{\mydata};
	\pgfmathtruncatemacro{\PlotNum}{\PlotNum-1}
	\addplot+[yshift=\pgflinewidth,mark options={fill=red!50!white}] table[x=datasets,y=lars]{\mydata};
	\pgfmathtruncatemacro{\PlotNum}{\PlotNum-1}
	\addplot +[yshift=\pgflinewidth,mark options={fill=brown!50!white}] table[x=datasets,y=HSIC-Lasso]{\mydata};
	\pgfmathtruncatemacro{\PlotNum}{\PlotNum-1}
	\addplot+[yshift=\pgflinewidth,mark options={fill=black!50!white}] table[x=datasets,y=Fast]{\mydata};
	\pgfmathtruncatemacro{\PlotNum}{\PlotNum-1}
	\addplot +[yshift=\pgflinewidth,color=tyrianpurple!50!black ,mark options={fill=tyrianpurple!50!white}] table[x=datasets,y=Saola]{\mydata};
	\pgfmathtruncatemacro{\PlotNum}{\PlotNum-1}
	\addplot+[yshift=\pgflinewidth,color=laserlemon!50!black,solid ,mark options={fill=laserlemon!50!white}] table[x=datasets,y=CCM]{\mydata};
	\pgfmathtruncatemacro{\PlotNum}{\PlotNum-1}
	\addplot +[yshift=\pgflinewidth,pink!50!black,solid,mark options={fill=pink!50!white}] table[x=datasets,y=BCOA]{\mydata};
	\pgfmathtruncatemacro{\PlotNum}{\PlotNum-1}
	\addplot +[yshift=\pgflinewidth,green!50!black,solid,mark options={fill=green!50!white}] table[x=datasets,y=DRPT]{\mydata};
	\end{axis}
	\end{tikzpicture}
		
	}
	\caption{Classification accuracies (CA) based on SVM  using mRMR, LARS, HSIC-Lasso, Fast-OSFS,  group-SAOLA, CCM, BCOA and DRPT over 10 runs for different number of features}
	\label{genomicresults}
\end{figure}
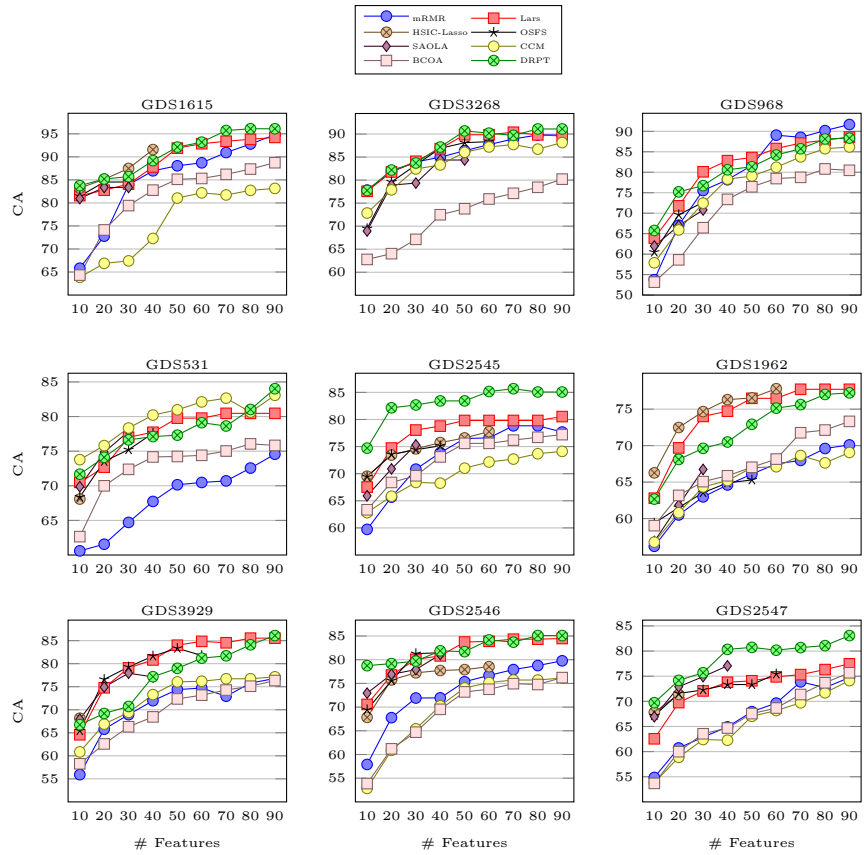


	We also take advantage of  IBM\textsuperscript \textregistered LSF  to report  running time,  CPU time and memory usage of each FS method.  We just remark that  through parallelization, an algorithm might achieve a better running time at the cost of having greater  CPU time. 
	
\begin{figure}[]
		\begin{center}
			\begin{tikzpicture} 	
			\begin{axis}[
			width  = 1*\textwidth,
			height = 7 cm,
			major x tick style = transparent,
			ybar=2*\pgflinewidth,
			bar width=2.5pt,
			ymajorgrids = true,
			ylabel = {Running Time (Second)},
			ymode=log,
			log basis y={10},
			symbolic x coords={GDS1615,GDS3268,GDS968,GDS531,GDS2545,GDS1962,GDS3929,GDS2546,GDS2547,NX100,NX200},
			x tick label style={rotate=45, anchor=north east, inner sep=0mm},
			tick label style={font=\tiny},
			xtick=data,
			scaled y ticks = false,
			enlarge x limits=0.06,
			ymin=0,
			legend columns=4,
			legend cell align=left,
			legend style={font=\fontsize{1}{1}\selectfont,at={(1,1)},anchor=north east}
			]
			\addplot table[x=datasets,y=mRMR]{\mydata};
			\addplot table[x=datasets,y=LARS]{\mydata};
			\addplot table[x=datasets,y=HSIC-Lasoo]{\mydata};
			\addplot table[x=datasets,y=Fast-OSFS]{\mydata};
			\addplot table[x=datasets,y=group-SAOLA]{\mydata};
			\addplot [color=pink!50!black,fill=pink!50!white] table[x=datasets,y=BCOA]{\mydata};
			\addplot [color=green!50!black,fill=green!50!white] table[x=datasets,y=DRPT]{\mydata};
			\legend{mRMR,LARS,HSIC-Lasoo,Fast-OSFS,group-SAOLA,BCOA,DRPT}

			\end{axis}
			\end{tikzpicture}
		\end{center}
		\caption{Running Time of feature selection by DRPT, HSIC-Lasso, LARS,  Fast-OSFS,  group-SAOLA, and mRMR over 10 runs using SVM}
		\label{Runtime}
	\end{figure}
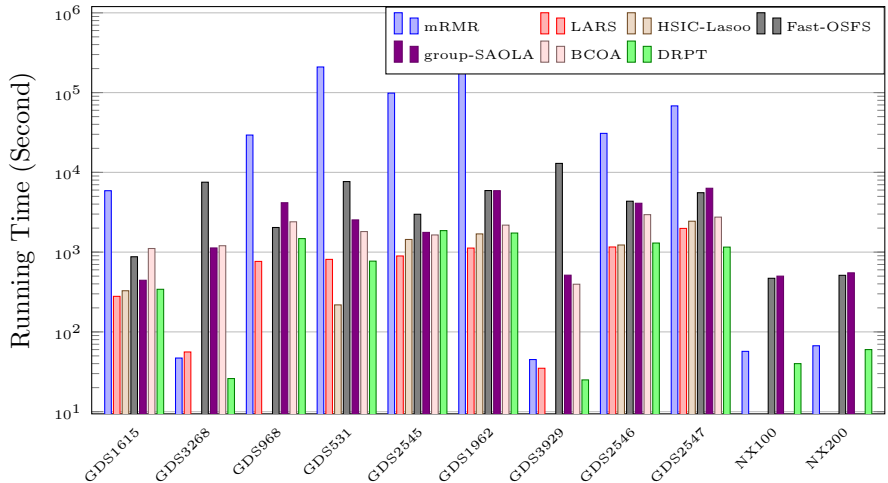

	Figure \ref{Runtime} depicts running time of FS methods that includes the classification time using SVM as well. We can see that LARS, HSIC-Lasso, and DRPT have comparable running times. The running times of  Fast-OSFS, group-SAOLA and BCOA are higher than DRPT  while the running time of mRMR is the worst among all by order of magnitude.

	\pgfplotstableread[row sep=\\,col sep=&]{
		datasets & mRMR & LARS & HSIC-Lasoo & Fast-OSFS &group-SAOLA&  BCOA&DRPT \\
		GDS1615  & 5888 & 279  & 328  		&874.944982	&444	                &	1110   &342\\
		GDS3268  & 47   & 56   &    		&7528.410845	&1128	                &	1203   &	26\\
		GDS968 	 &29363 & 765  &     		&2036.107312	&4161	             &	2400    &1478\\
		GDS531   &209523& 811  & 218  		&7668.470718	&2535	         &	1811    &	771\\
		GDS2545  & 98509& 895  &  1441 		&2983.530370	&1768	        &	1642   &	1861\\
		GDS1962  &449641& 1124 & 1692  		&5916.406140	&5892	      &	2180     &	1735\\
		GDS3929  & 45   & 35   &    		&12952.274760	&514	                &	396     &	25\\
		GDS2546  & 30822& 1160 & 1228  		&4347.832412	& 4100	       &	2950  &	1300\\
		GDS2547  & 68153& 1983 & 2441  		& 5558.892699	& 6321	       &	2747  &1156\\
		NX100    & 57   &      &    		&470								&500			&				&	40\\
		NX200    & 67   &      &    		&512								&550			&				&	60\\
	}\mydata

	Next, we compare CPU time.  For a non-parallelized algorithm, the CPU time is almost the as same as the running time.  However,  a parallelized  algorithm takes more  CPU time as it hires multi-processes. Figure \ref{cpu} shows the CPU time that is taken by FS methods on six  common datasets.  Clearly, mRMR   takes the  highest CPU time and it is also obvious that HSIC-Lasso uses more processes as it is implemented in parallel. 
	
	We also quantify the computational performance of all methods based on the peak memory usage over six common datasets (Figure \ref{mem}). We observe that mRMR and HSIC-Lasso require
	an order of magnitude higher memory than LARS. Although  the peak memory usage by DRPT is significantly  lower than 
	mRMR and HSIC-Lasso, DRPT takes almost the same amount of memory across all datasets. In this regard, there is a potential for  more efficient implementation of DRPT.

	\pgfplotstableread[row sep=\\,col sep=&]{
		datasets & mRMR & LARS & HSIC-Lasoo & Fast-OSFS& group-SAOLA&DRPT \\
		GDS1615  & 6500 & 357  & 2597  		&950	&512	&480\\
		GDS531   &210000& 916  & 3914  		&7850	&1328	&	850\\
		GDS2545  & 99500& 958  &  3648		&3148	&2064	&	1739\\
		GDS1962  &894911& 1328 & 10267 		&6213	&6326	&	1925\\
		GDS2546  & 31022& 1260 & 3826  		&4486	&4469	&	1463\\
		GDS2547  & 68500& 1295 & 42			&	5856	&	6610	&1765\\
	}\cpudata

	\pgfplotsset{width=7cm,compat=1.9}
	\usepgfplotslibrary{statistics}
	
	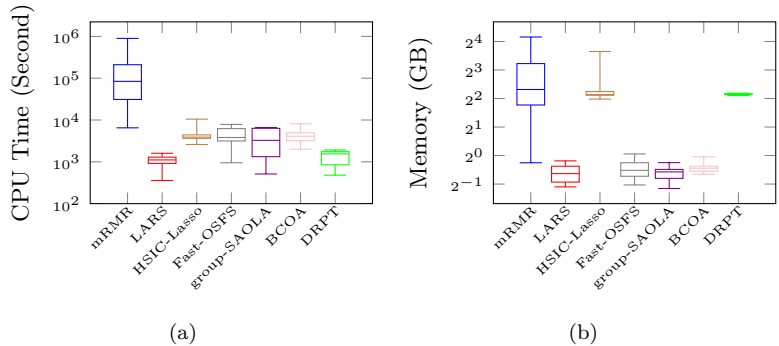
\begin{figure}[t]
		\centering 
		\subfigure[]{
			\begin{tikzpicture}
			\begin{axis}[
			width  = 0.44*\textwidth,
			height = 4cm,
			boxplot/draw direction=y,
			ylabel style={font=\small},
			ylabel = {CPU Time (Second)},
			ymode=log,
			log basis y={10},
			xtick={1,2,3,4,5,6,7},
			ytick={100,1000,10000,100000,1000000},
			xticklabels={mRMR,LARS,HSIC-Lasso,Fast-OSFS,group-SAOLA,BCOA,DRPT},
			x tick label style={rotate=45, anchor=north east, inner sep=0mm},
			tick label style={font=\tiny},
			ymin=100,
			]
			\addplot [blue,boxplot prepared={
				lower whisker=6500, lower quartile=31022,
				median=84000, upper quartile=210000,
				upper whisker=894911},
			] coordinates {};
			\addplot [red,boxplot prepared={
				lower whisker=357, lower quartile=916,
				median=1110, upper quartile=1295,
				upper whisker=1600},
			] coordinates {};
			\addplot [brown,boxplot prepared={
				lower whisker=2597, lower quartile=3648,
				median=3900, upper quartile=4400,
				upper whisker=10500},
			] coordinates {};
			\addplot [gray,boxplot prepared={
				lower whisker=950, lower quartile=3148,
				median=3817, upper quartile=6213,
				upper whisker=7850},
			] coordinates {};
			\addplot [violet,boxplot prepared={
				lower whisker=512, lower quartile=1328,
				median=3266.5, upper quartile=6326,
				upper whisker=6610},
			] coordinates {};
			\addplot [pink,boxplot prepared={
				lower whisker=2000, lower quartile=3192,
				median=4089, upper quartile=4958,
				upper whisker=8100},
			] coordinates {};
			\addplot [green,boxplot prepared={
				lower whisker=480, lower quartile=850,
				median=1550, upper quartile=1750,
				upper whisker=1925},
			] coordinates {};
			
			\end{axis}
			\end{tikzpicture}
			\label{cpu}}
		\subfigure[]{
			\begin{tikzpicture}
			\begin{axis}[
			width  = 0.44*\textwidth,
			height = 4cm,
			boxplot/draw direction=y,
			ylabel style={font=\small},
			ylabel = {Memory (GB)},
			ymode=log,
			log basis y={2},
		    xtick={1,2,3,4,5,6,7},
			xticklabels={mRMR,LARS,HSIC-Lasso,Fast-OSFS,group-SAOLA,BCOA,DRPT},
			x tick label style={rotate=45, anchor=north east, inner sep=0mm},
			tick label style={font=\tiny},
			ytick={0.5,1,2,4,8,16,32},
			ymin=0,
			]
			\addplot [blue,boxplot prepared={
				lower whisker=859/1024, lower quartile=3500/1024,
				median=5100/1024, upper quartile=9579/1024,
				upper whisker=18284/1024},
			] coordinates {};
			\addplot [red,boxplot prepared={
				lower whisker=478/1024, lower quartile=537/1024,
				median=660/1024, upper quartile=790/1024,
				upper whisker=900/1024},
			] coordinates {};
			\addplot [brown,boxplot prepared={
				lower whisker=4040/1024, lower quartile=4437/1024,
				median=4500/1024, upper quartile=4850/1024,
				upper whisker=12827/1024},
			] coordinates {};
			\addplot [gray,boxplot prepared={
				lower whisker=501/1024, lower quartile=619/1024,
				median=715/1024, upper quartile=860/1024,
				upper whisker=1065/1024},
			] coordinates {};
			\addplot [violet,boxplot prepared={
				lower whisker=460/1024, lower quartile=588/1024,
				median=688/1024, upper quartile=731/1024,
				upper whisker=862/1024},
			] coordinates {};
			\addplot [pink,boxplot prepared={
				lower whisker=650/1024, lower quartile=700/1024,
				median=753/1024, upper quartile=790/1024,
				upper whisker=995/1024},
			] coordinates {};
			\addplot [green,boxplot prepared={
				lower whisker=4400/1024, lower quartile=4500/1024,
				median=4550/1024, upper quartile=4598/1024,
				upper whisker=4650/1024},
			] coordinates {};
			\end{axis}\label{Mem}
			\end{tikzpicture}
			\label{mem}}
		\caption{(a) CPU Time and (b) Memory taken by DRPT, HSIC-Lasso , LARS,  Fast-OSFS,  group-SAOLA, and mRMR}
		\label{cpu_mem}
	\end{figure}

	We had to leave the CCM method out of the comparison in Figures \ref{Runtime} and \ref{cpu_mem}  {because it is implemented in Python and required a high volume of RAM while the other methods implemented in Matlab. Table \ref{ccm_results} shows the CCM performance in terms of running time, CPU time, and memory usage, where running time and CPU time are measured by second and memory usage is scaled in GB.

	\begin{table}[]
			\centering
			\caption{ Running time, CPU time and memory taken by CCM model}
		\begin{tabular}{lccc}
			\hline
			Dataset & \multicolumn{1}{l}{Running Time} & \multicolumn{1}{l}{CPU Time} & \multicolumn{1}{l}{Memory} \\ \hline
			GDS1615 & 850                              & 26855                        & 107                        \\
			GDS531  & 3327                             & 36193                        & 150                        \\
			GDS2545 & 1478                             & 36009                        & 74                         \\
			GDS1962 & 3621                             & 38730                        & 148                        \\
			GDS2546 & 1389                             & 35331                        & 73                         \\
			GDS2547 & 985                              & 30500                        & 71                         \\ \hline
		\end{tabular}
			\label{ccm_results}
	\end{table}

	\section{Conclusions}\label{Conclusions}
	
	In this paper, we presented a linear  feature selection method  (DRPT) for high-dimensional genomic datasets. The novelty of our method is to remove irrelevant features outright and then detect correlations on the reduced dataset using perturbation theory.  
	While we showed DRPT precisely  detects irrelevant and redundant features on a synthetic dataset, the extent to which DRPT is effective on real dataset was tested on ten genomic datasets. 
	We demonstrated  that  DRPT performs well on these datasets compared to state-of-the-art feature selection algorithms. We proved that DRPT is robust against noise.  Performance of DRPT is insensitive to permutation of rows or columns of the data. Even though the running time of DRPT is comparable  to other FS methods,  an efficient implementation of DRPT in Python or \CC	can help improve both  memory usage and  the running time.
	
	In this paper, we focused only on genomic datasets because inherently they are similar. For example, they all have full-row rank. Besides, it is widely
	accepted that there is no dimension reduction algorithm that performs well on all datasets (compared to other methods). In a future work, we aim to revise our current algorithm to offer a new FS algorithm that performs well on face and text datasets.
 
 \section*{Acknowledgements}
 The authors would like to thank the anonymous reviewers for  valuable and practical comments  that  helped  to improved the paper. The research is supported by  NSERC of Canada under grant \#RGPIN 418201.

\bibliographystyle{elsarticle-num}

\end{document}